\documentclass[twoside]{article}
\usepackage[accepted]{aistats2018}
\usepackage[letterpaper, pass]{geometry}

\usepackage{times}
\usepackage{hyperref}      
\usepackage{url}            
\usepackage{amsfonts}       
\usepackage{algorithm}
\usepackage{algorithmic}
\usepackage{xcolor}
\usepackage{amsmath}
\usepackage{amsthm}
\usepackage{amssymb}
\usepackage{tikz}
\usepackage{etoolbox}
\usepackage{subcaption}
\usepackage{multicol}
\usepackage{comment}
\newcommand{\smallcirc}[1]{\scalebox{#1}{$\circ$}}

\DeclareMathOperator*{\wsum}{%
\mathchoice%
  {\ooalign{\phantom{$\displaystyle\sum$}\cr\hidewidth\raisebox{1.2\height}{$\mkern22mu\smallcirc{0.7}$}\hidewidth\cr%
                                  \raisebox{-0.7\height}{$\mkern22mu\smallcirc{0.7}$}\cr
                                  \hidewidth$\displaystyle\sum$}}
  {\ooalign{$\textstyle\sum$\cr%
                                \hidewidth\raisebox{1.9\height}{$\mkern16mu\smallcirc{0.4}$}\hidewidth\cr
                                \hidewidth\raisebox{-.3\height}{$\mkern16mu\smallcirc{0.4}$}\hidewidth\cr}}
  {\ooalign{\raisebox{0\height}{\scalebox{.6}{$\scriptstyle\sum$}}\cr%
                                \hidewidth\raisebox{1.6\height}{$\mkern7.5mu\smallcirc{0.2}$}\hidewidth\cr
                                \hidewidth\raisebox{-0.2\height}{$\mkern7.5mu\smallcirc{0.2}$}\hidewidth\cr}}
  {\ooalign{\raisebox{.2\height}{\scalebox{.6}{$\scriptstyle\sum$}}\cr%
                                \hidewidth\raisebox{2.2\height}{$\mkern7.5mu\smallcirc{0.2}$}\hidewidth\cr
                                \hidewidth\raisebox{0.4\height}{$\mkern7.5mu\smallcirc{0.2}$}\hidewidth\cr}}
}

\newtheorem{theorem}{Theorem}

\newtheorem{example}{Example}

\newcommand{\eff}{\mathnormal{f}}

\begin{document}

\twocolumn[

\aistatstitle{Gauged Mini-Bucket Elimination for Approximate Inference}

\aistatsauthor{ Sungsoo Ahn \And Michael Chertkov \And  Jinwoo Shin \And Adrian Weller }

\aistatsaddress{ Korea Advanced Institute \\ of Science and Technology
\And Los Alamos \\ National Laboratory,\\
Skolkovo Institute of \\ Science and Technology
\And Korea Advanced Institute \\ of Science and Technology
\And University of Cambridge, \\
The Alan Turing Institute}


]

\begin{abstract}
Computing the partition function $Z$ of a discrete graphical model is a fundamental inference challenge. 
Since this is computationally intractable, variational
approximations  
are often used 
in practice. 
Recently, so-called gauge transformations were used to improve
variational 
lower bounds on $Z$.
In this paper, we propose a new gauge-variational approach, termed WMBE-G, 
which combines gauge transformations 
with the weighted mini-bucket elimination (WMBE) method.
WMBE-G can provide both upper and lower bounds
on $Z$, and is easier to optimize than the prior gauge-variational algorithm. 
We show that WMBE-G strictly improves the earlier WMBE approximation for symmetric models
 including Ising models with no magnetic field. 
Our experimental results demonstrate the effectiveness of WMBE-G even for generic, nonsymmetric models. 
\end{abstract}

\section{INTRODUCTION}

Graphical Models (GMs) express the 
factorization of the joint multivariate probability distribution over subsets of variables 
via graphical relations among them. 
GMs have been developed 
in information theory \cite{gallager1962low,kschischang1998iterative},
physics \cite{35Bet,36Pei,87MPZ,parisi1988statistical,09MM}, artificial intelligence \cite{pearl2014probabilistic}, and machine learning \cite{jordan1998learning,freeman2000learning}.
For a GM,
computing the partition function $Z$ (the normalization constant)
is a fundamental 
inference task of great interest.
However, this 
task is known to be computationally intractable in general:
it is \#P-hard even to approximate
\cite{jerrum1993polynomial}.

Variational 
approaches frame the inference task
as an optimization problem, which is typically solved approximately. 
Key challenges for variational methods 
are to scale efficiently with the number of variables; and to try to provide guaranteed upper or lower bounds on $Z$.  

Popular variational methods include: 
the mean-field (MF) approximation \cite{parisi1988statistical}, which provides a lower bound on $Z$; the tree-reweighted (TRW) approximation \cite{wainwright2005new}, which provides an upper bound; 
and belief propagation (BP) \cite{pearl1982reverend},
which often performs well but provides neither an upper nor lower bound in general. 
Other variational methods have been investigated for providing lower bounds
\cite{liu2010negative, ermon2012density, liu2011bounding, novikov2014putting}
or upper bounds
\cite{ermon2012density,liu2011bounding, novikov2014putting} for approximating $Z$.

Methods using \emph{reparametrizations} \cite{03WJW}, \emph{gauge transformations} (GT) \cite{06CCa,06CCb} 
or \emph{holographic transformations} (HT) \cite{valiant2008holographic,al2011normal} have been explored. 
These methods each 
consider modifying the base GM by transforming the potential factors in various ways, aiming to simplify the inference task, 
while keeping the partition function $Z$ unchanged. 
We call these methods collectively $Z$-\emph{invariant methods}.  
See \cite{08JW,forney2011partition,Misha_notes} for discussions of the differences and relations between these methods.

An approach to combine variational and $Z$-invariant methods was recently introduced by \cite{ahn2017gauge}, yielding a lower bound on $Z$. 
They proposed gauge-variational optimization formulations 
built upon MF and BP, incorporating  
the generic IPOPT solver \cite{wachter2006implementation} as an essential inner optimization routine. 
Here we introduce a new  
gauge-variational optimization approach,
using variational methods other than MF and BP, and employing a specialized solver for inner optimization which is more efficient than IPOPT. Further, our approach yields lower and upper bounds on $Z$. 


\noindent {\bf Contribution.}
We develop a new family of gauge-variational algorithms
combining the methods of gauge transformations (GTs) and weighted mini-bucket elimination (WMBE) \cite{liu2011bounding}.
The significance of our new approach, which we call WMBE-G, is twofold: 
\begin{itemize}
\item[$\mathcal C1.$] 
We introduce optimization formulations which provide both upper and lower bounds of $Z$
by generalizing the original WMBE bounds to incorporate GTs.
The authors \cite{liu2011bounding} use 
the re-parameterization framework, which is a distribution-invariant method
that is a strict sub-class of GTs. 
Hence, our formulations 
explore a strictly larger freedom in optimization, which we observe typically leads to significantly better bounds in practice.  
Indeed, we 
provide an analytic class of GMs (symmetric binary GMs including Ising models with no magnetic field)
where ours provide strictly better results. 
\item[$\mathcal C2.$] 
We propose a novel optimization solver alternating between 
gauges and factors to minimize (or maximize) the proposed objectives, and demonstrate its computational advantages. 
We remark that
the earlier optimization approaches in \cite{ahn2017gauge}
required `non-negativity' constraints which are tricky to handle, while we do not.
\cite{ahn2017gauge} addresses the challenge using the generic IPOPT solver with the log-barrier method, 
but it is not clear if this will scale well for large instances. 
On the other hand, our proposed algorithms are clearly scalable since
they solve purely unconstrained optimizations in a distributed manner.
\end{itemize}

Our experimental results show that WMBE-G has superior performance in comparison with other known algorithms, including WMBE.
We remark that the main contribution of WMBE \cite{liu2011bounding}
was to introduce H\"older weights to improve
the original mini-bucket elimination (BE) bound 
\cite{dechter2003mini},
whereas we additionally optimize gauges for even better performance.
In our experiments, we observe that
the contribution of H\"older weights
is relatively marginal compared to gauges in optimizing the BE bound (see Section \ref{sec:exp} for more details). 
Namely, we found that gauges are more crucial than H\"older weights for better approximation to $Z$,
while the computational costs of optimizing them are similar.
In this paper, we mainly focus on WMBE-G using the H\"older inequality 
to obtain
an upper bound on $Z$, but a lower bound can be similarly derived
using the reverse H\"older inequality (see Section \ref{sec:wmbe}).

\section{PRELIMINARIES}
\subsection{Graphical Models}

{\bf Factor-graph GM.}
We consider an undirected, bipartite factor graph 
$G=(V,E)$ with vertices $V=X \cup F$ comprising variables $X$ and factors $F$, 
and edges between variable and factor nodes $E\subseteq X\times F$.  
Each random variable 
$x_{v} \in X$ is discrete, taking values in $\{1,\cdots,d\}$.  
The distribution factorizes as follows:
\begin{equation}\label{eq:factorgraphgm}
p(\mathbf{x}) = \frac{1}{Z} \prod_{\alpha \in F}\eff_{\alpha}(\mathbf{x}_{\alpha}).
\end{equation}
Here, $\mathcal{F} = \{\eff_{\alpha}\}_{\alpha\in F}$ 
is a set of non-negative
functions called \emph{factors}, 
and 
$\mathbf{x}_{\alpha}$ is the subset of variables for 
factor $\alpha$, 
i.e., $\mathbf{x}_{\alpha} = [x_{v}: v \in N(\alpha)]$ 
with $N(\alpha) = \{v:(v,\alpha)\in E\}$. 
The normalization constant
$$Z:= \sum_{\mathbf{x}}
  \prod_{\alpha\in F} \eff_{\alpha}(\mathbf{x}_{\alpha})$$
is called the {\it partition function}. 
It is well known that the partition function
is computationally intractable in general:  it is \#P-hard 
even to approximate  
\cite{jerrum1993polynomial}. 

\begin{figure}[t!]
\centering
    \includegraphics[width=0.22\textwidth]{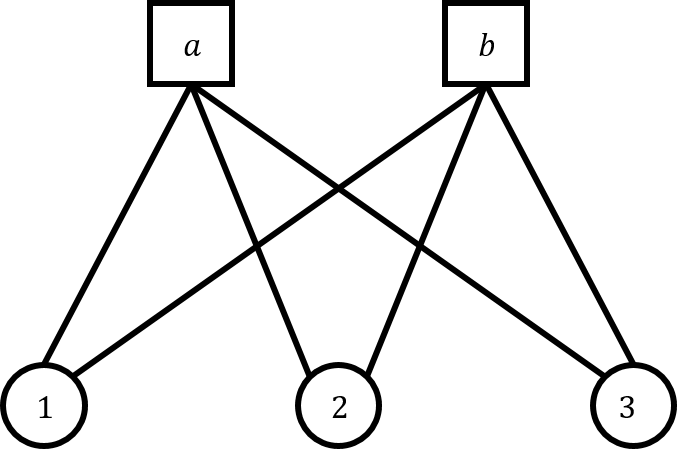}
    \hspace{0.1in}
    \includegraphics[width=0.22\textwidth]{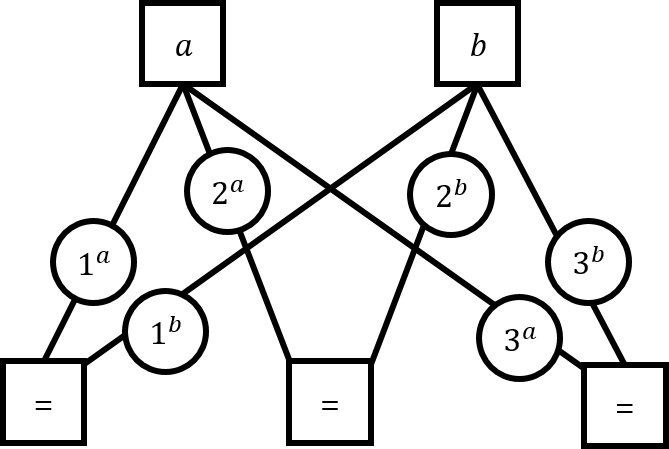}
    \caption{
    {Example of transformation from the factor-graph GM (left) to the Forney-style GM (right). 
    Squares and circles indicate factors and variables respectively. 
    New factors denoted as `=' force adjoining variables be consistent, i.e., have the same value. 
    }
    }
    \label{fig:factor2forney}
\end{figure}

{\bf Forney-style GM.}
For ease of notation in dealing with GTs, 
throughout this paper
we shall assume Forney-style GMs \cite{forney2001codes}. These ensure  
that every 
variable  
has two 
adjacent factors, i.e., $|N(v)|=2 ~\forall~v\in X$. 
As shown in \cite{06CCa,06CCb},
Forney-style GMs provide a more compact description of gauge transformations
without any loss of generality: 
given any factor-graph GM,
one can construct an equivalent Forney-style GM \cite{levin2007tensor}. See Figure \ref{fig:factor2forney} for an example.

\subsection{Gauge Transformations}\label{subsec:GT}
Gauge transformations \cite{06CCa,06CCb}
are a family of linear transformations of 
the factor functions in \eqref{eq:factorgraphgm}
which leave the
the partition function $Z$ 
invariant. 
GTs are defined by
the following set of invertible 
$d \times d$ matrices 
$\{G_{v\alpha}: 
(v, \alpha)\in E\}$, 
termed \emph{gauges}:
\begin{equation*}
  G_{v\alpha}
  = \left[\begin{array}{ccc}
  G_{v\alpha}(1,1) & \cdots & G_{v\alpha}(1,d)\\
  \vdots & \ddots & \vdots\\
  G_{v\alpha}(d,1) & \cdots & G_{v\alpha}(d,d)\\
  \end{array}
  \right].
\end{equation*}
The transformed GM
with respect to the gauges 
$\mathbf{G} = \{G_{v\alpha}:(v,\alpha) \in E\}$
consists of modified factors $\{\widehat{\eff}_{\alpha}: \alpha \in F\}$ computed as follows:
\begin{equation}
\label{eq:gaugetransformation}
\widehat{\eff}_{\alpha}(\mathbf{x}_{\alpha};\mathbf{G}_{\alpha})
= \sum_{\mathbf{x}_{\alpha}^\prime}
\eff_{\alpha}(\mathbf{x}_{\alpha}^\prime)
\prod_{v \in N(\alpha)} G_{v\alpha}(x_{v}, x^\prime_{v}),
\end{equation}
where $\mathbf{G}_{\alpha} = \{G_{v\alpha}:v\in N(\alpha)\}$. 
Here, the gauges must satisfy 
the following {\it gauge constraints}:
\begin{equation}
\label{eq:gaugeconstraint}
G_{v\alpha}^{\top} G_{v\beta} = \mathbb{I},\qquad 
\forall v \in X,
\end{equation}
where $\mathbb{I}$ is the identity matrix and 
$N(v)=\{\alpha,\beta\}$ (recall that we 
assume $|N(v)| = 2$). 
With these constraints, the partition function is known to be 
invariant under the transformation \cite{06CCa,06CCb}, i.e.,
\begin{equation*}
Z ~=~ \sum_{\mathbf{x}}
\prod_{\alpha\in F} \eff_{\alpha}(\mathbf{x}_{\alpha})
~=~ \sum_{\mathbf{x}}
\prod_{\alpha \in F} \widehat{\eff}_{\alpha}(\mathbf{x}_{\alpha};
\mathbf{G}_{\alpha}).
\end{equation*} 
Thus gauges lead to the transformed
distribution
    $p(\mathbf{x};\mathbf{G}) =  \prod_{\alpha\in F}\widehat{\eff}_{\alpha}(\mathbf{x}_{\alpha};\mathbf{G}_{\alpha})/Z$.
We remark that it might be invalid when 
$\widehat{\eff}_{\alpha}(\mathbf{x}_{\alpha};\mathbf{G}_{\alpha})$ is negative.
Nevertheless, even in this case, the partition function invariance still holds.
We provide an example of a gauge transformation in the Supplement.
\subsection{Weighted Mini-Bucket Elimination}\label{sec:wmbe}

{\it Bucket (or variable) elimination} (BE) 
\cite{dechter1999bucket, koller2009probabilistic} is 
a method for computing the partition function exactly
based on directly summing out the variables sequentially. 
First, BE assumes a
fixed elimination ordering 
$o = [v_{1},\cdots, v_{n}]$ 
among variables nodes 
$v\in X$. 
Then BE groups factors by placing each 
factor $\eff_{\alpha}$ in the 
``bucket'' $B_{v}$ of its 
earliest argument $v\in N(\alpha)$ 
appearing in the elimination order $o$.
Next, BE eliminates the variable by 
marginalizing the product of factors in the bucket, 
i.e., 
\begin{equation}
\label{eq:BE}
    \eff_{B_{v}}(\mathbf{x}_{B_{v}}) = 
    \sum_{x_{v}}\prod_{\eff_{\alpha}\in B_{v}}\eff_{\alpha}(\mathbf{x}_{\alpha}) 
    \qquad \forall~\mathbf{x}_{B_{v}},
\end{equation}
where $\mathbf{x}_{B_{v}} = [x_{u}:u\in var(B_{v}), u \neq v]$ 
and $var(B_{v})$ 
indicates the subset of variables in the bucket. 
Finally, the newly generated function $\eff_{B_{v}}$ 
is inserted into another bucket 
corresponding to its earliest argument in the 
elimination order. 
This process is easily seen as 
applying a distributive property: 
groups of factors corresponding to buckets
are summed out sequentially, 
and then the newly created factor 
(without the eliminated variable)
is assigned to another bucket.

The computational cost of BE is 
exponential in the number of 
uneliminated variables in the bucket, i.e., 
the 
{\it induced width}\footnote{The minimum possible
induced with is called {\it tree-width}.} of the graph 
given the elimination order.
BE is 
summarized in Algorithm \ref{alg:bucket}.

\begin{algorithm}[h!]
\caption{BE for computing $Z$}
\label{alg:bucket}
\begin{algorithmic}[1]
\STATE {\bf Input:} GM on graph $G=(V,E)$ with $V = (X,F)$ and factors $\mathcal{F} = \{\eff_{\alpha}\}_{\alpha\in F}$ 
and elimination order $o=[v_{1},\cdots,v_{n}]$.
\vspace{0.05in}
  \hrule
    \vspace{0.05in}
\STATE $\mathcal{F}^\prime \leftarrow \mathcal{F}$
\FOR{$v$ in $o$}
\STATE $B_{v} \leftarrow \{\eff_{\alpha}|\eff_{\alpha}\in \mathcal{F}, v \in N(\alpha)\}$
\STATE Generate new factor by:
\begin{equation*}
\eff_{B_{v}}(\mathbf{x}_{B_{v}}) = 
    \sum_{x_{v}}\prod_{\eff_{\alpha}\in B_{v}}\eff_{\alpha}(\mathbf{x}_{\alpha}), ~\forall~ \mathbf{x}_{B_{v}}.
    \end{equation*}
\STATE $\mathcal{F}^\prime \leftarrow \mathcal{F}^\prime \cup \{\eff_{B_{v}}\} - B_{v}$
\ENDFOR
\vspace{0.05in}
  \hrule
  \vspace{0.05in}
\STATE {\bf Output:} $Z=\prod_{\eff_{\alpha}\in \mathcal{F}^\prime}\eff_{\alpha}$
\end{algorithmic}
\end{algorithm}

{\it Mini-bucket elimination} (MBE) \cite{dechter2003mini} and
{\it weighted mini-bucket elimination} (WMBE) \cite{liu2011bounding} 
approximate BE
by splitting computation of each 
bucket into several ``mini-buckets'', where WMBE
additionally makes use of 
H\"older's inequality \cite{hardy1934inequalities}.
Since 
MBE is a special
case of WMBE (by choosing extreme H\"older weights), 
here we focus on providing 
background for WMBE.

Let $\{\psi_{i}(x), i=1,\cdots, m\}$ be some functions 
defined on discrete variable $x$, and
$\mathbf{w} = [w_{1},\cdots w_{n}]$ be 
a vector of {\it H\"older weights}. We define a 
{\it weighted absolute summation}, defined as follows: 
\begin{equation*}
    \wsum^{w_{i}}_{x}\psi_{i}(x) := \big(\sum_{x} |\psi_{i}(x)|^{1/w_{i}}\big)^{w_{i}}.
\end{equation*}
Equivalently, $\wsum^{w_{i}}_{x}$ is the Schatten $p$-norm 
with $p=1/w_{i}$. 
If $w_{i} > 0$ for all $i\geq 1$, 
then H\"older's inequality implies that 
\begin{equation}
\label{eq:holder}
    \wsum^{w_{0}}_{x}\prod_{i=1}^{m}\psi_{i}(x) \leq \prod_{i=1}^{m}\wsum^{w_{i}}_{x}\psi_{i}(x),
\end{equation}
where $w_{0}=\sum_{i}w_{i}$.
If only one weight is positive, 
e.g., $w_{1}>0$ and 
$w_{i}<0$ for all $i >1$, 
we have the reverse H\"older's inequality:
\begin{equation}
    \wsum^{w_{0}}_{x}\prod_{i=1}^{m}\psi_{i}(x) \geq \prod_{i=1}^{m}\wsum^{w_{i}}_{x}\psi_{i}(x).
\end{equation}

\begin{algorithm}[!ht]
\caption{WMBE for bounding $Z$}
\label{alg:weightedminibucket}
\begin{algorithmic}[1]
\STATE {\bf Input:} GM on graph $G=(V,E)$ 
with $V=(X,F)$, 
factors $\mathcal{F} = \{\eff_{\alpha}\}_{\alpha\in F}$,
elimination order $o=[v_{1},\cdots,v_{n}]$ and bound on 
bucket size $ibound$.
\vspace{0.05in}
  \hrule
\vspace{0.05in}
\STATE $\mathcal{F}^{\prime}\leftarrow \mathcal{F}$
\FOR{$v$ in $o$}
\STATE $B_{v} \leftarrow \{\eff_{\alpha}|\eff_{\alpha}\in \mathcal{F}^{\prime}, v \in \partial \alpha\}$
\STATE Partition $B_{v}$ into $R_{v}$ subgroups 
$\{B_{v}^{r}\}_{r=1}^{R_{v}}$ 
such that $|var(B_{v}^{r})| \leq ibound$ for all $r$.
\STATE Assign weights $w_{1},\cdots,w_{R_{v}}$ while satisfying $\sum_{r}w_{r} = 1$.
\FOR{$r \leftarrow1,\cdots,R_{v}$ }
\STATE Generate a new factor by:
\begin{equation*}
\eff_{B_{v}^{r}}(\mathbf{x}_{B_{v}^{r}}) =  
    \wsum_{x_{v}}^{w_{r}}\prod_{\eff_{\alpha}\in B_{v}^{r}}\eff_{\alpha}(\mathbf{x}_{\alpha}), ~\forall~ \mathbf{x}_{B_{v}}.
    \end{equation*}
\STATE $\mathcal{F}^{\prime} \leftarrow \mathcal{F}^{\prime} \cup \{\eff_{B^{r}_{v}}\} - B_{v}^{r}$
\ENDFOR
\ENDFOR
\vspace{0.05in}
  \hrule
\vspace{0.05in}
\STATE {\bf Output:} $Z_{\text{WMBE}}=\prod_{\eff_{\alpha}\in \mathcal{F}^{\prime}}\eff_{\alpha}$
\end{algorithmic}
\end{algorithm}

WMBE modifies BE by  
applying  H\"older's inequality whenever 
the size of a bucket, i.e., 
length of $\mathbf{x}_{B_{v}}$, 
exceeds some given parameter called 
$ibound$. 
In this case, 
WMBE splits the 
bucket into multiple 
`mini-buckets', and 
weighted absolute summation 
is evaluated sequentially in
place of \eqref{eq:BE}, i.e., 
\begin{equation*}
\sum_{x_{v}}\prod_{\eff_{\alpha}\in B}|\eff_{\alpha}(\mathbf{x}_{\alpha})|\leq 
\prod_{r=1}^{R_{v}}\wsum_{x_{v}}^{w_{r}}\prod_{\eff_{\alpha}\in B_{v}^{r}}\eff_{\alpha}(\mathbf{x}_{\alpha}), 
\end{equation*}
for all $\mathbf{x}_{B_{v}}$, 
where H\"older weights satisfy $\sum_{r}w_{r}=1 , w_{r} > 0$, 
$B_{v} = \bigcup_{r}B_{v}^{r}$, and $B_{v}^{r}$ is disjoint for all $r$.
We then generate multiple new factors by:
\begin{equation*}
    \eff_{B_{v}^{r}}(\mathbf{x}_{B_{v}^{r}}) =
    \wsum_{x_{v}}^{w_{r}}\prod_{\eff_{\alpha}\in B_{v}^{r}}\eff_{\alpha}(\mathbf{x}_{\alpha}), \qquad \forall~ \mathbf{x}_{B_{v}},
\end{equation*}
and insert into other buckets.
By construction, WMBE
yields an upper bound for 
the partition function $Z$. 
One can use the same idea to derive a lower bound for $Z$
using the reverse H\"older's inequality. 
We summarize WMBE in 
Algorithm \ref{alg:weightedminibucket}.

One can interpret MBE as a special case
of WMBE by 
assigning a single weight to be close to $1$ 
and 
others to be close to $0$, i.e., 
$\mathbf{w}=\lim_{w\rightarrow 
0^{+}}[1-w, w, w,\cdots]$.
Instead, Liu and Ihler \cite{liu2011bounding}  
optimize the H\"older weights so that
WMBE can outperform MBE, 
 which we discuss further
in Section \ref{sec:Gauged}.

\section{GAUGED WMBE ALGORITHM}\label{sec:Gauged}
In this section, we describe our 
gauge optimization scheme WMBE-G to improve the previous WMBE bound, 
yielding  
gauranteed upper bound approximations for 
the partition function $Z$. 
Our scheme improves the 
standard WMBE bound by searching over the large family of gauge transformed (possibly invalid) GMs 
to find the tightest WMBE bound possible. 

\subsection{Key Optimization Formulation}
In order to describe the optimization formulation 
for tightening the WMBE bound, we 
first 
observe that 
\eqref{eq:WMBEbound} 
can be reformulated into 
\begin{equation}
\label{eq:holdermodified}
    \sum_{x_{v}}\prod_{\eff_{\alpha}\in B}|\eff_{\alpha}(\mathbf{x}_{\alpha})| 
    \leq 
    \sum_{x_{v}^{(1)}}^{w_{1}}\cdots\wsum_{x_{v}^{(R_{v})}}^{w_{R_{v}}}
    \prod_{r=1}^{R_{v}}
    \prod_{\alpha \in B_{v}^{r}}
    \eff_{\alpha}(\mathbf{x}_{\alpha\backslash v}, x_{v}^{(r)})
\end{equation}
where $\mathbf{x}_{\alpha\backslash v} = [x_{u}:u\in N(\alpha), u\neq v]$.
While notation is  complex, this is simply applying the 
distributive property on the right hand side of 
\eqref{eq:WMBEbound}.
The procedure can 
be seen as `splitting'
variable from $x_{v}$ to $x_{v}^{(1)},\cdots x_{v}^{(R_{v})}$ and 
its associated node from $v$ to $v^{(1)},\cdots v^{(R_{v})}$ 
so that factors no longer share the split variable. 
We remark that under Forney-style GMs, $R_{v}\leq2$ since 
exactly $2$ factors are associated with a variable.
After repeatedly applying the inequality, we arrive at the following WMBE bound, termed {\it weighted partition function}:
\begin{equation}
\label{eq:WMBEbound}
    Z \leq Z_{\text{WMBE}}=\wsum^{\bar{w}_{\bar{n}}}_{\bar{x}_{\bar{1}}}\cdots\wsum^{\bar{w}_{1}}_{\bar{x}_{1}}
    \prod_{\alpha \in F}\eff_{\alpha}(\bar{\mathbf{x}}_{\alpha}).
\end{equation}
In \eqref{eq:WMBEbound}, $\bar{\mathbf{x}} = 
[\bar{x}_{1},\cdots, \bar{x}_{\bar{n}}]$ 
and $\bar{\mathbf{w}} = 
[\bar{w}_{1},\cdots, \bar{x}_{\bar{n}}]$
indicate the `split' version of variables 
and associated H\"older weights, 
indexed by appearance of associated node in the
modified elimination order 
$\bar{o} = [v_{1}^{(1)},\cdots v_{1}^{(R_{v_{1}})},\cdots ,v_{n}^{(1)},\cdots v_{n}^{(R_{v_{n}})}]$.
Therefore, the WMBE bound can be seen as a 
weighted absolute summation over product of factors in a new GM. 
However, unlike the original partition function, 
the weighted absolute summation 
is 
tractable with respect to $ibound$  since 
at most $d^{ibound}$ terms are counted
for each weighted absolute summation, 
or equivalently 
variable elimination of mini-buckets.
Finally, we are able to present our main optimization formulation:
\begin{align}
\label{eq:WMBEG}
    \operatornamewithlimits{\mbox{minimize}}_{\mathbf{G}} 
    \quad
    &\wsum^{\bar{w}_{\bar{n}}}_{\bar{x}_{\bar{n}}}\cdots\wsum^{\bar{w}_{1}}_{\bar{x}_{1}}
    \prod_{\alpha \in F}\widehat{\eff}_{\alpha}(\bar{\mathbf{x}}_{\alpha};\mathbf{G}_{\alpha}), \\
    \mbox{subject to}\quad
    &G_{v\alpha}^{\top}G_{v\beta} = \mathbb{I}, \quad
    \forall ~ v \in X, N(v) = \{\alpha,\beta\}. \notag 
\end{align} 

\subsection{Algorithm Description} 
\label{subsec:alg}

We now describe 
an efficient algorithm to optimize \eqref{eq:WMBEG}. 
First, the gauge constraint can be removed simply by  
expressing one (of the two) gauges in terms of the other, 
e.g., $G_{v\beta}$ via 
$(G_{v\alpha}^{\top})^{-1}$. 
Then, \eqref{eq:WMBEG} can be optimized via 
any type of unconstrained optimization solver. 
Here, we optimize gauges by 
gradient descent followed 
by additional updates on factor values. 

To this end, we initialize gauges by identity matrices, 
which immediately yields the original WMBE bound from \eqref{eq:WMBEbound} 
since 
$\eff_{\alpha}(\mathbf{x}_{\alpha}) = 
\widehat{\eff}_{\alpha}(\mathbf{x}_{\alpha};\mathbb{I}_{\alpha})$,
where $\mathbb{I}_{\alpha} = [G_{v\alpha} = \mathbb{I}:(v,\alpha)\in E]$.
Next, under expressing gauges via one another, i.e., $G_{v\beta}\leftarrow \left(G_{v\alpha}^{\top}\right)^{-1}$, 
we 
update each gauge element by 
gradient descent
for minimization of the
weighted $\log$ partition function upper bound $\log Z_{\text{WMBE}}$ as follows:
\begin{align}
\label{eq:gaugeupdate1}
G_{v\alpha}(x_{v}^{\prime},x_{v}^{\prime\prime})~\leftarrow~ & G_{v\alpha}(x_{v}^{\prime},x_{v}^{\prime\prime}) - \mu \frac{\partial \log Z_{\text{WMBE}}}{\partial G_{v\alpha}(x_{v}^{\prime},x_{v}^{\prime\prime})}\notag \\ 
    \frac{\partial \log Z_{\text{WMBE}}}{\partial G_{v\alpha}(x_{v}^{\prime},x_{v}^{\prime\prime})}~ =~&
    \sum_{\bar{\mathbf{x}}_{\alpha\backslash v}}
    q(\bar{\mathbf{x}}_{\alpha\backslash v}, x_{v}^{\prime\prime})
    \frac{\eff_{\alpha}(\bar{\mathbf{x}}_{\alpha\backslash v}, x_{v}^{\prime})}
    {\eff_{\alpha}(\bar{\mathbf{x}}_{\alpha\backslash v}, x_{v}^{\prime\prime})}\notag\\
    & \- \!\!\!\!\!\! -\sum_{\bar{\mathbf{x}}_{\beta\backslash v}}
    q(\bar{\mathbf{x}}_{\beta\backslash v}, x_{v}^{\prime})
    \frac{\eff_{\beta}(\bar{\mathbf{x}}_{\beta\backslash v}, x_{v}^{\prime\prime})}
    {\eff_{\beta}(\bar{\mathbf{x}}_{\beta\backslash v}, x_{v}^{\prime})},
\end{align}
where $\mu$ is the step size\footnote{{See Section \ref{sec:exp} for details of our choice of step size.}}, 
$\mathbf{x}_{\alpha\backslash v} = [x_{u}:u\in N(\alpha), u\neq v]$
and $q$ is an `auxiliary distribution' defined as
\begin{align*}
    q(\bar{\mathbf{x}}) &= 
    \prod_{k=1}^{\bar{n}}q(\bar{x}_{k}|\bar{x}_{k+1:\bar{n}}),\\
    q(\bar{x}_{k}|\bar{x}_{k+1:n}) &\propto
    \left(\wsum_{\bar{x}_{k-1}}^{\bar{w}_{k-1}}\cdots \wsum_{\bar{x}_{1}}^{\bar{w}_{1}}
    \prod_{\alpha \in F}\eff_{\alpha}(\bar{\mathbf{x}}_{\alpha})\right)^{1/\bar{w}_{k}}. 
\end{align*} 
We also update $G_{v\beta}\leftarrow (G_{v\alpha}^{\top})^{-1}$
and the value of associated factors by the gauge-transformed factors, i.e., 
\begin{equation}
\label{eq:gaugeupdate2}
\eff_{\alpha}(\mathbf{x}_{\alpha})\leftarrow \widehat{\eff}_{\alpha}(\mathbf{x}_{\alpha};
\mathbf{G}_{\alpha}),
\end{equation} 
and similarly for $\eff_{\beta}$.
Finally, for the next iteration, we reset $G_{v\alpha}\leftarrow \mathbb{I}$.

The above update leads to
an improved WMBE bound, which can be repeated
for 
better bounds (until convergence).
Each iteration $t=1,\dots T$ results in a sequence of gauges $\mathbf{G}^{(t)}$ 
obtained by \eqref{eq:gaugeupdate1}, 
and factors $\eff^{(t)}_{\alpha}$ obtained by \eqref{eq:gaugeupdate2} 
can be expressed as
$\eff^{(t)}_{\alpha}(\mathbf{x}_{\alpha}) = \widehat{\eff}^{(0)}_{\alpha}(\mathbf{x}_{\alpha} ; \mathbf{G}_{\alpha}^{\prime}), 
$
where $\eff^{(0)}_{\alpha}=\eff_{\alpha}$ is the original GM factor, and 
$\mathbf{G}_{\alpha}^{\prime}$ consists 
of gauges $G_{v\alpha}^{\prime} =
G_{v\alpha}^{(t+1)}G_{v\alpha}^{(t)}\cdots 
G_{v\alpha}^{(1)}$ for $v \in N(\alpha)$. 
We remark that one can use  
na\"ive gradient descent, i.e.,
update gauges only (without resetting to identity matrices),
instead of factors as in \eqref{eq:gaugeupdate2}.
However, by utilizing the additional factor updates, 
the gradient formulation is simplified and 
redundant computations of gauge transformations are reduced. 
We summarize the above update procedure  
 in Algorithm \ref{alg:WMBEG}.

Furthermore, one can utilize 
ideas from \cite{liu2011bounding} 
in order to improve the efficiency 
and power of the proposed optimization. 
First,  
computation of auxiliary marginals 
$q(\mathbf{x}_{\alpha})$ in 
\eqref{eq:gaugeupdate1} can be 
efficiently carried out by a
message-passing 
scheme proposed by the authors.
Moreover, one can 
jointly optimize 
the H\"older weights 
$\bar{\mathbf{w}}$ in addition to $\mathbf{G}$ 
using the auxiliary distribution 
during optimization of \eqref{eq:WMBEG}. 
In our experiments, we utilize  
both the message-passing algorithm and 
the joint optimization involving $\bar{\mathbf{w}}$
using the log-gradient step proposed by the authors.

Finally, we remark that the elimination order and 
bucket split strategy might be another 
freedom that one may exploit in order to 
tighten the WMBE bound. However, 
their optimizations are hard (see \cite{liu2011bounding}).
Hence, 
we choose the elimination order arbitrarily 
in our experiments.
For the bucket split strategy,
if one assumes Forney-style GMs, 
any strategy reduces into 
a fixed split process, 
i.e., whenever $ibound$ is 
exceeded, a variable $x$ is 
always split in two parts $x^{(1)},x^{(2)}$, and 
adjacent factors are assigned 
separately. 


\begin{algorithm}[H]
\caption{Gauged WMBE for bounding $Z$}
\label{alg:WMBEG}
\begin{algorithmic}[1]
\STATE {\bf Input:} GM on graph $G=(V,E)$ 
with $V=(X,F)$, 
factors $\mathcal{F} = \{\eff_{\alpha}\}_{\alpha\in F}$,
elimination order $o=[v_{1},\cdots,v_{n}]$ and bound on 
bucket size $ibound$.
\vspace{0.05in}
  \hrule
\vspace{0.05in}
\STATE $\mathcal{F}^{\prime}\leftarrow \mathcal{F}$
\STATE $\bar{o}\leftarrow \emptyset, \bar{w} \leftarrow \emptyset$.
\STATE Initialize by $\bar{o} = \emptyset, \bar{\mathbf{w}} = \emptyset$.
\FOR{$v$ in $o$}
\STATE $B_{v}\leftarrow \{\eff_{\alpha}|\eff_{\alpha} \in \mathcal{F}^{\prime}, 
v \in N(\alpha)\}$
\STATE Partition $B_{v}$ into $R_{v}$ subgroups 
$\{B_{v}^{r}\}_{r=1}^{R_{v}}$ 
such that $|var(B_{v}^{r})| \leq ibound$ for all $r$. 
\STATE Assign weights $w_{1},\cdots,w_{R_{v}}$ while satisfying $\sum_{r}w_{r} = 1$.
\FOR{$r \leftarrow1,\cdots,R_{v}$ }
\STATE Generate a new factor by:
\begin{equation*}
\eff_{B_{v}^{r}}(\mathbf{x}_{B_{v}^{r}}) =  
    \wsum_{x_{v}}^{w_{r}}\prod_{\eff_{\alpha}\in B_{v}^{r}}\eff_{\alpha}(\mathbf{x}_{\alpha}), ~\forall~ \mathbf{x}_{B_{v}}.
    \end{equation*}
\STATE $\mathcal{F}^{\prime} \leftarrow \mathcal{F}^{\prime} \cup \{\eff_{B^{r}_{v}}\} - B_{v}^{r}$
\ENDFOR
\STATE Extend $\bar{o}$ by $[v^{(1)},\cdots, v^{(R_{v})}]$
\STATE Extend $\bar{\mathbf{w}}$ by $[w_{1},\cdots, w_{R_{v}}]$
\ENDFOR
\STATE Initialize by $G_{v\alpha} = \mathbb{I}$ for all $(v,\alpha) \in E$. 
\FOR{$t=1,2,\cdots ,T$}
\FOR{$v$ in $X$ with $N(v)=\{\alpha, \beta\}$}
\STATE Update $G_{v\alpha}$ by \eqref{eq:gaugeupdate1}.
\STATE $G_{v\beta}\leftarrow (G_{v\alpha}^{\top})^{-1}$
\STATE Set $\eff_{\alpha}(\mathbf{x}_{\alpha})\leftarrow \widehat{\eff}_{\alpha}(\mathbf{x}_{\alpha}; \mathbf{G}_{\alpha})$ and 
$
\eff_{\alpha}(\mathbf{x}_{\alpha})\leftarrow \widehat{\eff}_{\alpha}(\mathbf{x}_{\alpha}; \mathbf{G}_{\alpha})$ 
for all $\mathbf{x}_{\alpha}$, $\mathbf{x}_{\beta}$.
\STATE Reset gauges $G_{v\alpha}, G_{v\beta}\leftarrow \mathbb{I}$.
\ENDFOR
\ENDFOR
\vspace{0.05in}
  \hrule
\vspace{0.05in}
\STATE {\bf Output:} $Z_{\text{WMBE-G}}=\prod_{\eff^{\prime}\in \mathcal{F}^{\prime}}\eff$
\end{algorithmic}
\end{algorithm}
\vspace{-0.05in}

\subsection{Relation to Previous Work}
 H\"older's inequality holds even 
for negative-valued functions, so 
we do not need to put any additional 
constraint on non-negativity of factors, e.g., $\widehat{\eff}_{\alpha}(\bar{\mathbf{x}}_{\alpha};\mathbf{G}_{\alpha})\geq 0$.
Thus,  invalid gauged transformed GMs are allowed for \eqref{eq:WMBEG}.
This contrasts with the earlier work of \cite{ahn2017gauge}, where 
additional non-negativity constraints 
were needed to restrict the gauge transformations considered. 
Consequently, to our knowledge, our formulation is the first to
explore the full range of freedom in gauge transformations when combined with methods of variational inference for GMs.
Further, avoiding these non-negativity constraints simplifies our optimization procedure enabling an approach which scales much better than that of 
\cite{ahn2017gauge}.

We emphasize that our optimization 
formulation \eqref{eq:WMBEG} is a 
strict generalization of 
the approach of \cite{liu2011bounding} which
optimizes the WMBE bound with 
respect to {\it reparameterization} of GMs.
Specifically, 
the GM reparameterized 
with respect to reparameterization parameters 
$\pmb{\theta} = [\theta_{v\alpha}:(v,\alpha)\in E]$  
consists of factors: 
\begin{equation}
\label{eq:reparam}
\widehat{\eff}_{\alpha}(\mathbf{x}_{\alpha};\pmb{\theta}_{\alpha}) = 
\prod_{v \in N(\alpha)}\exp(\theta_{v\alpha}(x_{i}))\eff_{\alpha}(\mathbf{x}_{\alpha}), 
\end{equation}
where 
$\pmb{\theta}_{\alpha} = \{\theta_{v\alpha}:v\in N(\alpha)\}$.
Here, the reparameterization parameter $\theta_{v\alpha}$
is constrained 
to satisfy the following constraint: 
\begin{equation}
\label{eq:reparamconstraint}
\exp(\theta_{v\alpha}(x_{v}) + 
\theta_{v\beta}(x_{v})) = 1\quad\forall~v\in X, x_{v}, 
\end{equation} 
where $N(v) = \{\alpha,\beta\}$.
With this constraint, it is  
 easy to check that such transformations are 
distribution-invariant 
\cite{03WJW} 
and form a strict subset of gauge transformations.  
Alternatively,
when gauges are restricted to diagonal matrices with 
non-negative elements, \eqref{eq:gaugetransformation} and \eqref{eq:gaugeconstraint} match \eqref{eq:reparam} and \eqref{eq:reparamconstraint}, respectively.
Therefore, optimizing \eqref{eq:WMBEG}
is guaranteed to perform no worse than
that of \cite{liu2011bounding}. 
Formally, 
we provide the following analytic 
class of GMs 
where gauge transformations 
are expected to perform 
strictly better than 
reparameterizations.  
Here, we say
a function of binary variables is \emph{symmetric} 
if its value 
is invariant under 
a `flipping' of all variables in its scope, e.g., 
$\eff_{\alpha}(2,1,2)=
\eff_{\alpha}(1,2,1)$. 
\begin{theorem}
\label{thm:reparamopt} 
Consider a GM over binary variables (i.e., $d=2$) where every factor $\eff_{\alpha}$
is symmetric. 
Then, 
$\pmb{\theta}=\{\theta_{v\alpha}(x_{v})=0,~\forall(v,\alpha)\in E,x_{v}\}$ 
is always a solution of the following optimization:
\begin{align*}
    \operatornamewithlimits{\mbox{minimize}}_{\pmb{\theta}} 
    \quad
    &\wsum^{\bar{w}_{\bar{n}}}_{\bar{x}_{\bar{1}}}\cdots\wsum^{\bar{w}_{1}}_{\bar{x}_{1}}
    \prod_{\alpha \in F}\widehat{\eff}_{\alpha}(\bar{\mathbf{x}}_{\alpha};\pmb{\theta}_{\alpha}), \\
    \mbox{subject to}\quad
    &\exp(\theta_{v\alpha}(x_{v}) + 
\theta_{v\beta}(x_{v})) = 1
    \quad\forall~v\in X,x_{v}. 
\end{align*} 
\end{theorem}

The proof of Theorem \ref{thm:reparamopt} is given in the Supplement. 
It shows that
for symmetric GMs, 
e.g., the Ising model with no magnetic field,
reparameterization is impossible to  
improve the WMBE bound. 
On the other hand, gauges are expected to improve it
as we explain in what follows.
We first remark that 
the optimality condition for 
reparameterization 
is equivalent to the zero gradient 
condition for diagonal elements of gauges, 
i.e., $\sum_{\bar{\mathbf{x}}_{\alpha\backslash v}}q(\bar{\mathbf{x}}_{\alpha})=
\sum_{\bar{\mathbf{x}}_{\beta\backslash v}}q(\bar{\mathbf{x}}_{\beta})$, 
which aims to match the auxiliary marginals 
of variables split by WMBE. 
Under symmetric models, 
variables are 
indistinguishable from 
an auxiliary marginals point of view, which leads to Theorem \ref{thm:reparamopt}.
On the other hand, 
the 
zero gradient condition 
for non-diagonal gauges is harder to match since it 
takes local conditional dependency 
into account, e.g., considers
$\eff_{\alpha}(\bar{\mathbf{x}}_{\alpha\backslash v}, x_{v}^{\prime})/\eff_{\alpha}(\bar{\mathbf{x}}_{\alpha\backslash v}, x_{v}^{\prime\prime})$ upon evaluating the gradient. 
For symmetric GMs, the above reasoning for reparameterization
fails
since variables are 
distinguishable after conditioning, 
e.g., 
$\eff_{\alpha}(\bar{\mathbf{x}}_{\alpha\backslash v}, x_{v}^{\prime}) \neq 
\eff_{\alpha}(\bar{\mathbf{x}}_{\alpha\backslash v}, x_{v}^{\prime\prime})$. 
Namely, optimal gauges believably have non-diagonal elements.
Indeed, in all our experiments, 
gauge transformations 
significantly outperform reparameterizations.

\section{EXPERIMENTS}\label{sec:exp}

In this section, we report experimental results on 
performance of our proposed algorithms 
for the task of upper bounding the 
partition function $Z$. 
\begin{figure*}[t!]
\vspace{-0.05in}
\centering
 \begin{subfigure}[b]{.4\linewidth}
    \centering
    \includegraphics[width=.99\textwidth]{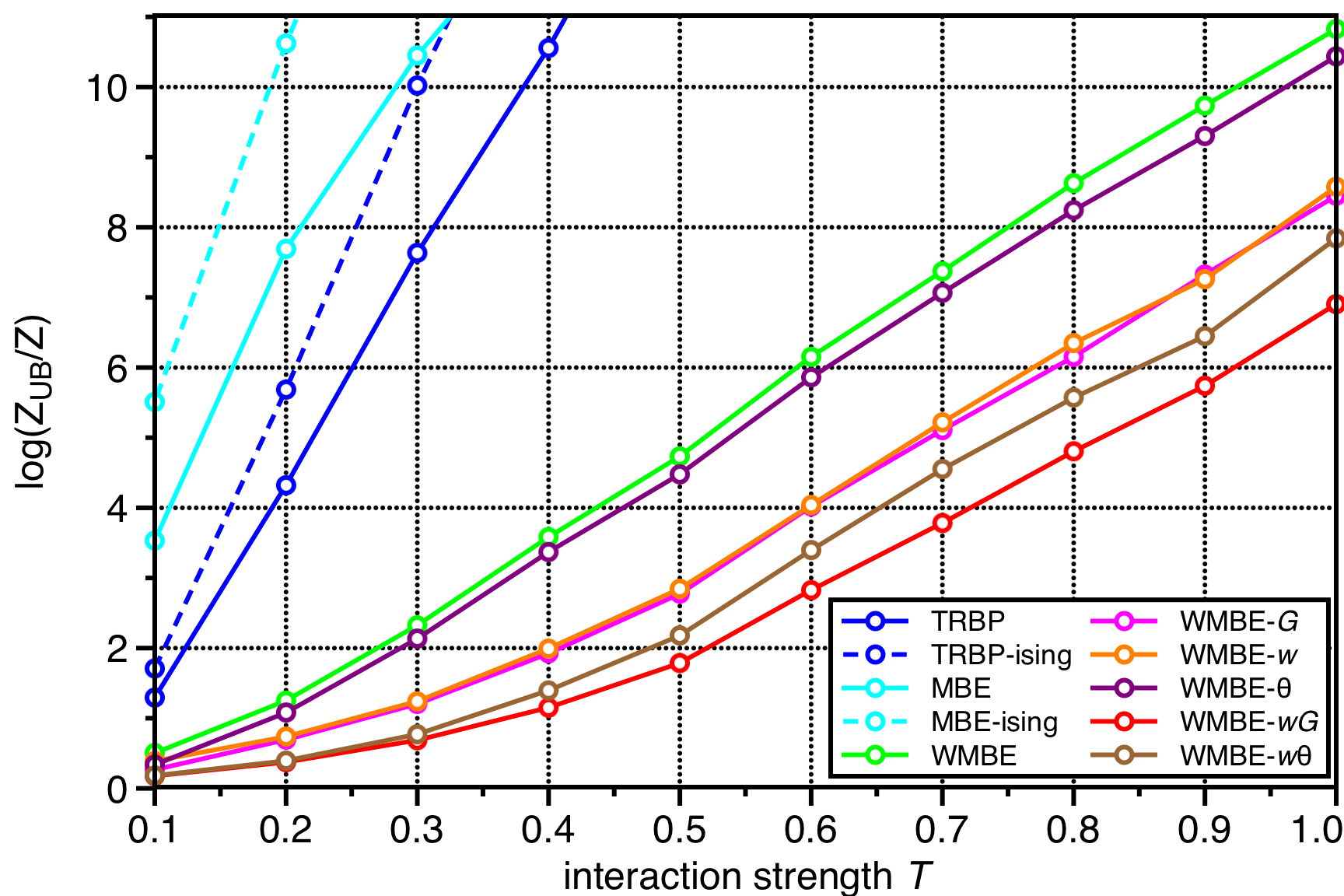}
    \caption{Ising grid GMs, $ibound=4$}
    \label{fig:ising1}
  \end{subfigure}
  \begin{subfigure}[b]{.4\linewidth}
    \centering
    \includegraphics[width=.99\textwidth]{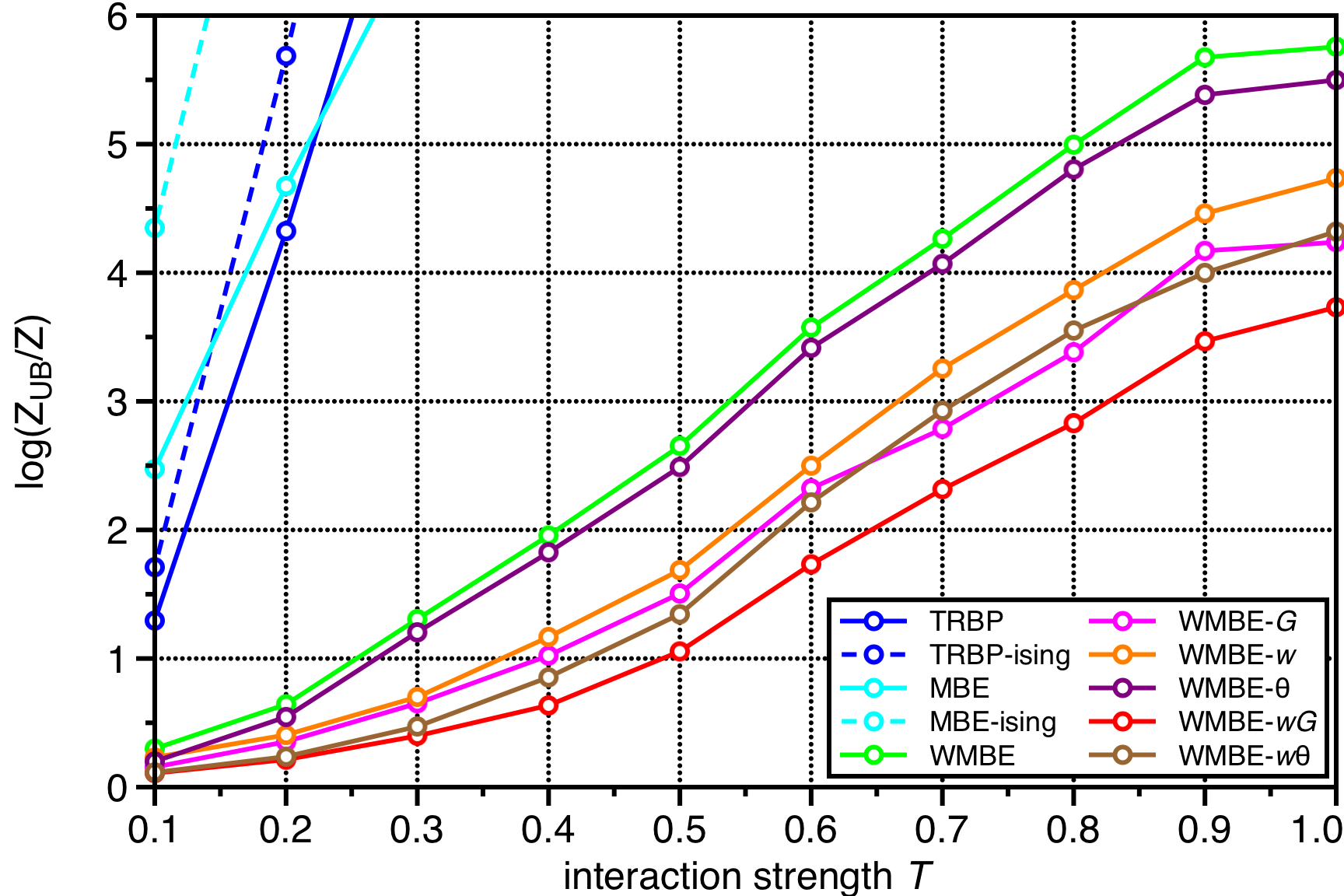}
    \caption{Ising grid GMs, $ibound=6$}
    \label{fig:ising2}
  \end{subfigure}
  \vspace{0.05in}\\
  \begin{subfigure}[b]{.3\linewidth}
    \centering
    \includegraphics[width=.99\textwidth]{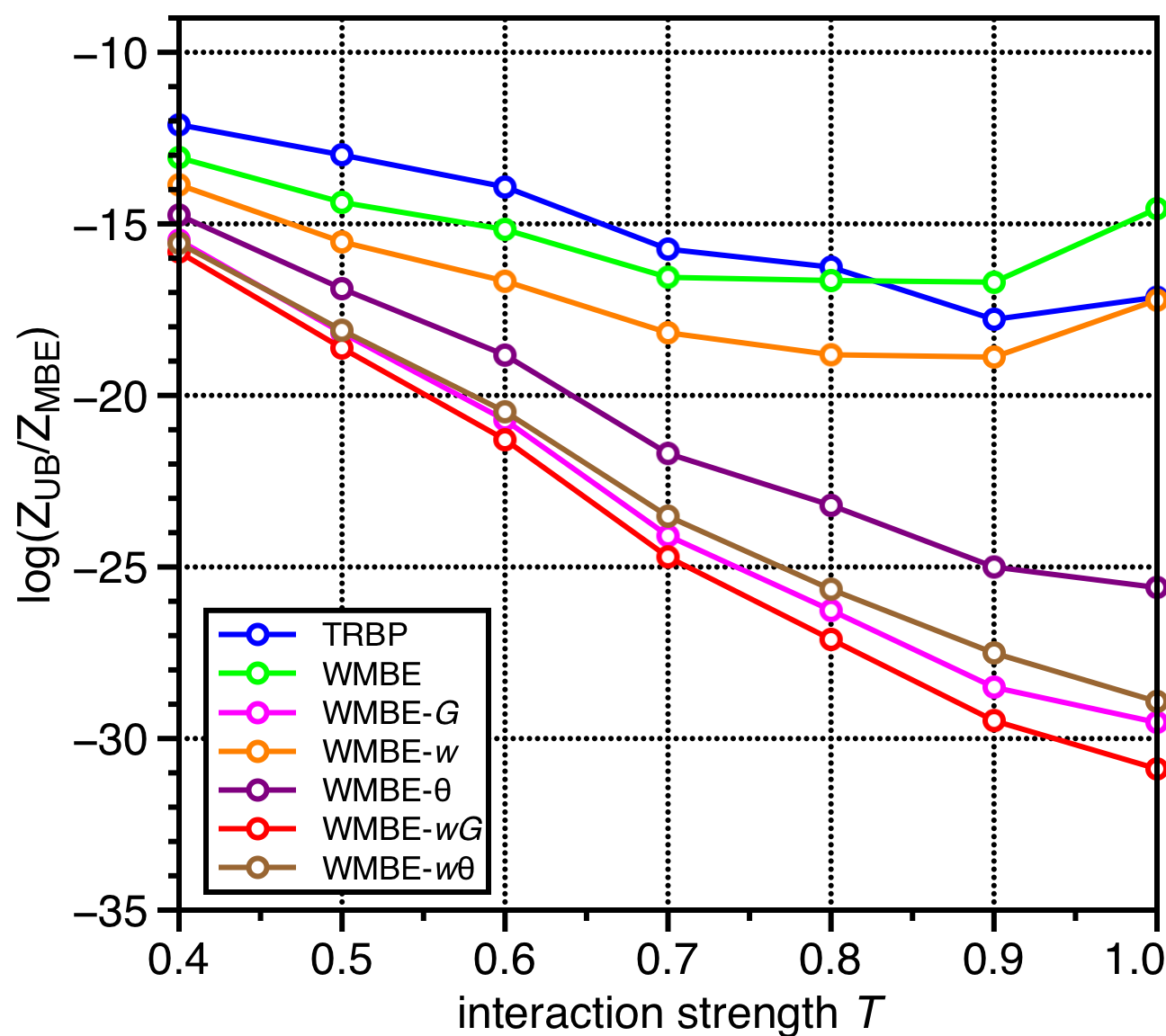}
    \caption{$3$-regular GMs, $ibound=4$}
    \label{fig:regular1}
  \end{subfigure}
  \begin{subfigure}[b]{.3\linewidth}
    \centering
    \includegraphics[width=.99\textwidth]{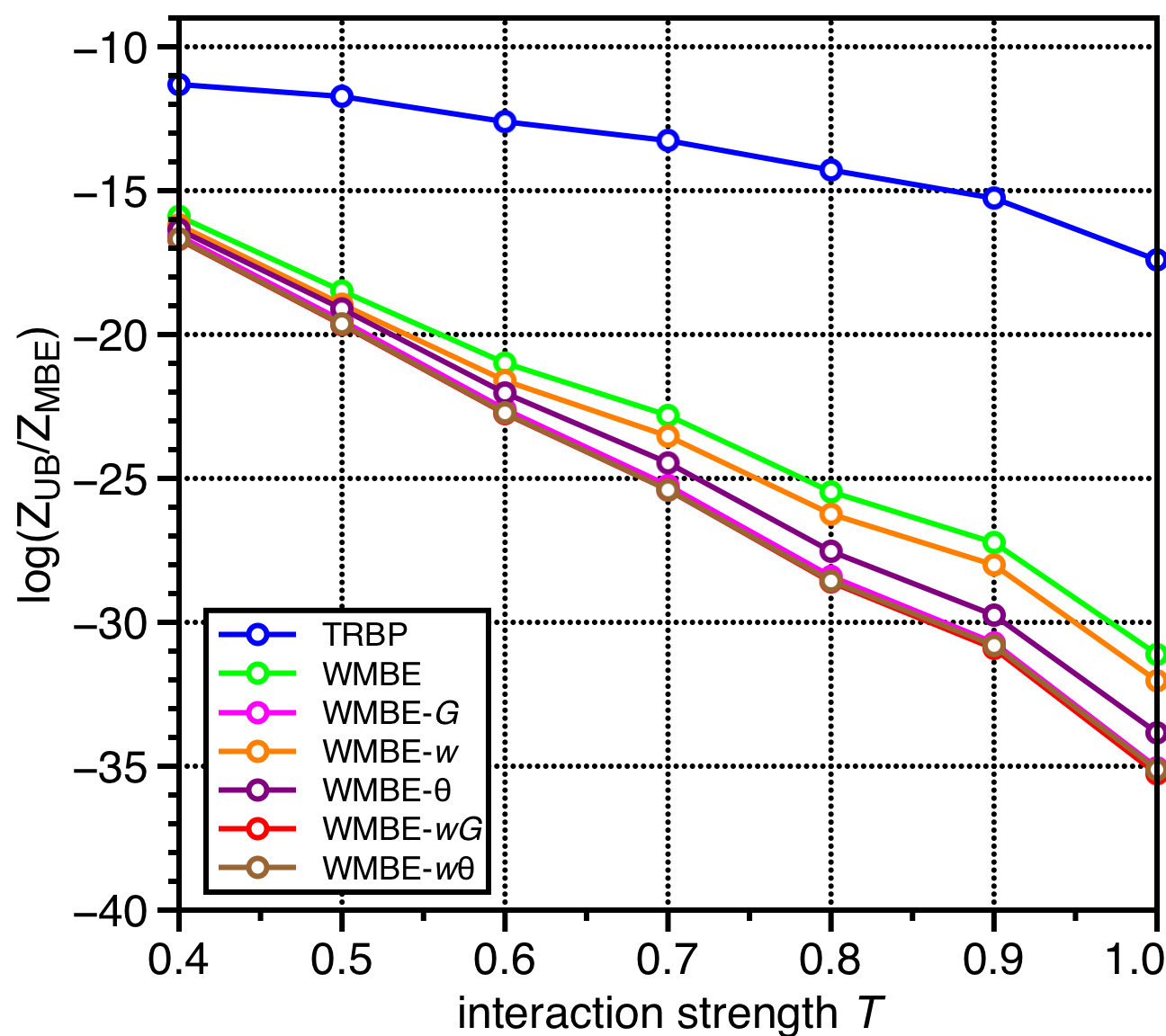}
    \caption{$3$-regular GMs, $ibound=6$}
    \label{fig:regular2}
  \end{subfigure}
  \begin{subfigure}[b]{.3\linewidth}
    \centering
    \includegraphics[width=.99\textwidth]{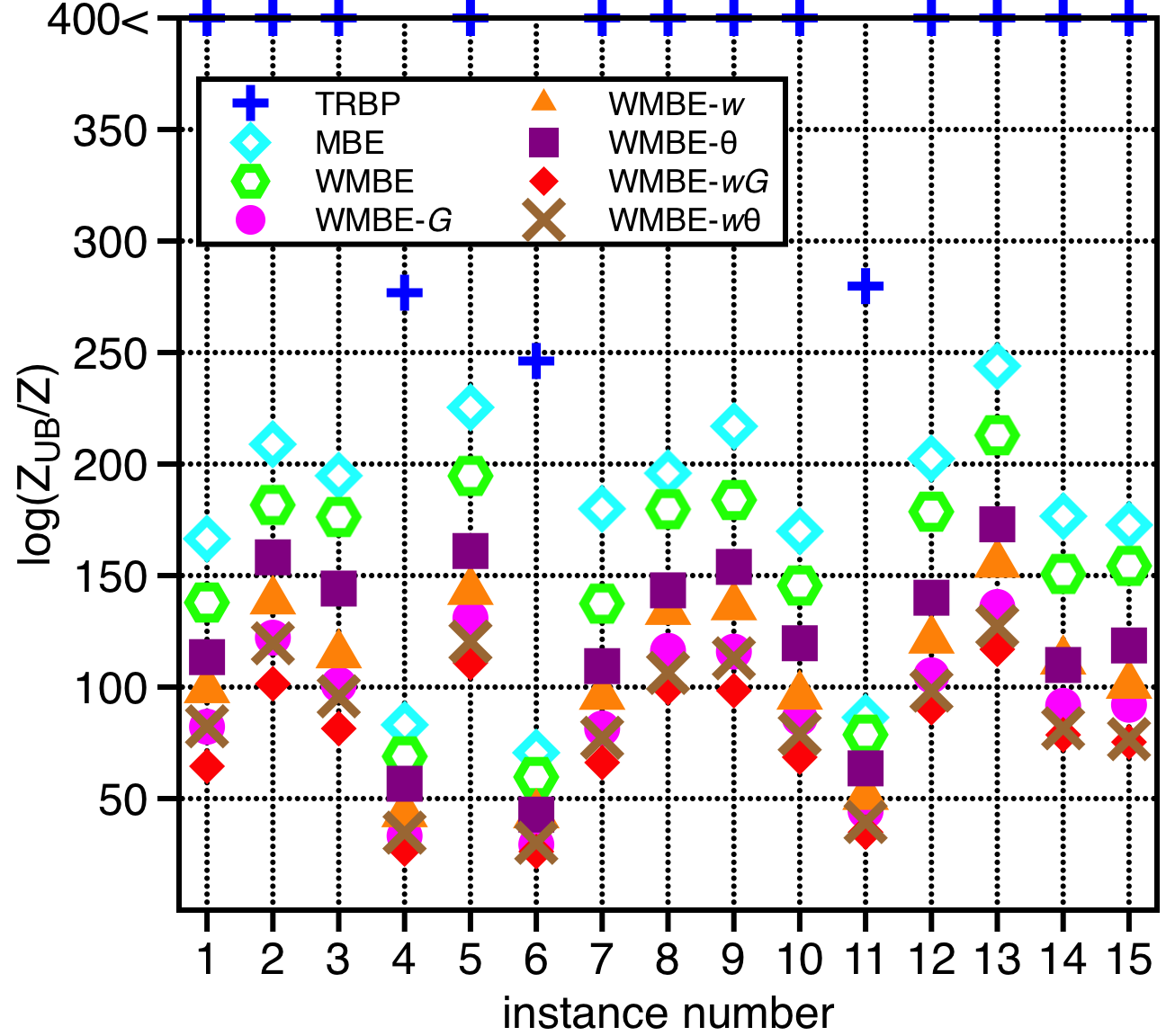}
    \caption{Linkage dataset, $ibound=6$.}
    \label{fig:uai}
\end{subfigure}
  \caption{{Performance comparisons in various families of GMs.}}
  \label{fig:performance}
 \vspace{-0.05in}
\end{figure*}

\subsection{Setup}
{Experiments were conducted with three family of GMs: 
(i) Ising models on a $10\times10$ grid graph (non-toroidal) with $180$ factors/$100$ variables; 
(ii) Forney-style GMs on the $3$-regular graph with $180$ factors/$270$ variables; 
and (iii) Linkage dataset from UAI 2014 Inference Competition \cite{uai}}. 

\begin{figure}[H]
\centering
    \includegraphics[width=.15\textwidth]{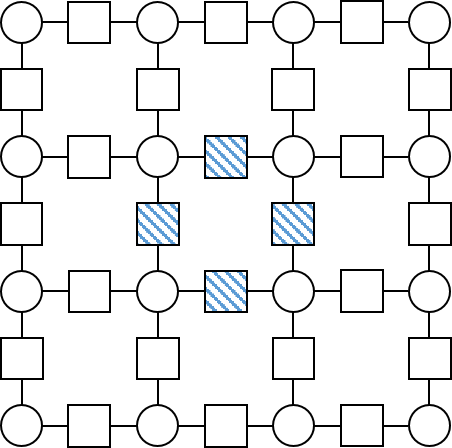}
    \hfill
    \includegraphics[width=.15\textwidth]{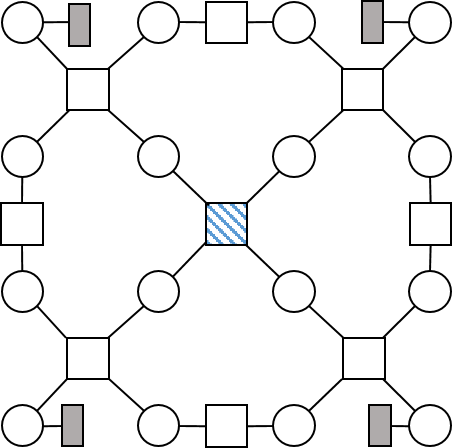}
    \hfill
    \includegraphics[width=.15\textwidth]{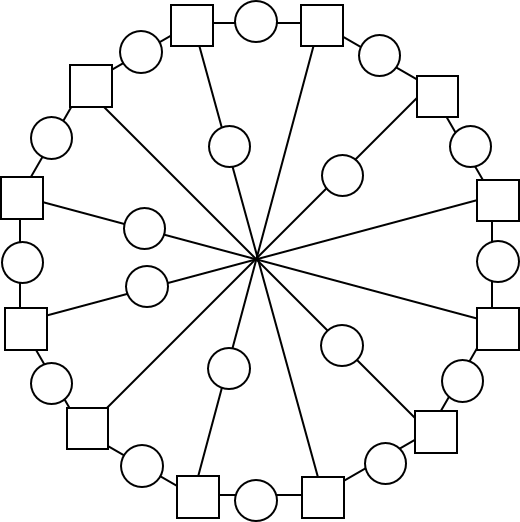}
  \caption{Illustration
  of Ising grid GM (left), its equivalent Forney-style GM (middle)  and 
  $3$-regular graph (right) of interest. 
  Factors surrounding the selected lattice (blue, dashed) are contracted into a single 
  factor, and then uniform single potentials (grey, filled) are added for variables of degree 1.}
  \label{fig:GMs}
  \vspace{-0.05in}
\end{figure}

\textbf{(i) Ising models. } 
Ising models were defined with mixed interactions (spin glasses):
\begin{equation*}  
    p(\mathbf{x}) = \frac{1}{Z}\exp\big(\sum_{v\in X}\phi_{v}x_{v} 
    + \sum_{(u,v)\in E}\phi_{uv}x_{u}x_{v}\big),
\end{equation*} 
where $x_{u}\in \{-1, 1\}$ and 
$\phi_{v}\sim\mathcal{N}(0,0.1)$, 
$\phi_{uv}\sim \mathcal{N}(0,T)$.
Here, $T \geq 0$ is the `interaction strength' parameter 
that controls the degree of interactions between variables.  
When $T=0$, variables are independent. As $T$ grows, the inference task is typically harder.

Note that this Ising model is 
not in Forney-style form, with  
variables adjacent 
to at most $4$ 
pairwise factors.
Hence, to apply our gauge optimization framework,
we generate an equivalent Forney-style GM 
using 
the transformation 
introduced in 
\cite{levin2007tensor}:
this maps
any classical lattice model (allowing for 
magnetic fields/singleton potentials) 
into an equivalent  
Forney-style model. 
At a high level, 
the transformation chooses disjoint lattices to 
cover the whole graph, 
then 
contracts each 
lattice into 
a single factor. 
Levin and Nave 
\cite{levin2007tensor} showed 
that one can always 
choose the lattice smartly so that 
each vertex is covered exactly twice,  
resulting in a Forney-style GM
(see Figure \ref{fig:GMs} for details).
Notably, this GM has 
relatively low induced width of $14$, 
thus the partition function 
can be computed exactly in reasonable time  
(though still computationally hard)
by using BE.

\textbf{(ii) 3-regular Forney-stlye GMs.} 
We considered $3$-regular Forney-style GMs 
with log-factors drawn from normal distribution, i.e.,
$\log\eff_{\alpha}(\mathbf{x}_{\alpha}) \sim \mathcal{N}(0, T)$. 
Again, $T\geq 0$ is the interaction strength parameter.
In this case, 
we would like to choose graphs so that
the induced width is high and the partition 
function is hard to compute.
To this end, 
we aligned 
factors in a cycle, 
and 
assigned variables (edges) 
between adjacent factors in the cycle
as well as those 
in the opposite side if it. 
See Figure \ref{fig:GMs} for its illustration.
This choice gives
high induced width,
e.g., na\"ively applying BE by 
eliminating variables between 
adjacent factors in  
clock-wise 
elimination order 
results in bucket 
size $2^{|V|/2 +2}$.

{
\textbf{(iii) UAI Linkage dataset.}
Finally, we consider 
a family of 
real-world models 
from the UAI 2014 
Inference Competition, 
namely the Linkage 
(genetic linkage) 
dataset. 
Specifically, the family 
consists of GMs with 
average of 
$949.94$ variables 
with averaged 
maximum cardinality  $\max_{i\in\mathcal{V}}|\mathcal{X}_{i}| = 4.95$ 
and $727.35$ 
non-singleton 
hyper-edges with 
averaged maximum 
size 
$\max_{\alpha\in\mathcal{E}}|\alpha|=4.47$. 
Since 
GMs in Linkage dataset 
were not of
Forney-style form, 
we constructed 
an equivalent 
Forney-style GM 
as in Figure 
\ref{fig:factor2forney}. 
}

{\bf Comparing approaches.}
We compared our gauged algorithm WMBE-G, i.e. 
optimizing the WMBE bound 
jointly with gauges and 
H\"older weights, to earlier methods considered in \cite{liu2011bounding}:  
the unoptimized WMBE bound (`WMBE'), 
its optimized versions with respect to 
H\"older weights $w$ and/or reparameterizations $\theta$ (`WMBE-$w$', 
`WMBE-$\theta$' and `WMBE-$w\theta$'). 
Further, we also ran the following popular baselines for computing upper bounds on $Z$: standard mini-bucket elimination 
(`MBE') and 
tree re-weighted belief propagation (`TRBP') \cite{wainwright2005new}.
Finally, for fair comparisons in Ising grid GMs,
we additionally compared to  
MBE and TRBP 
run 
on the original Ising grid GM 
(MBE-Ising and TRBP-Ising) 
in order to validate whether 
the forementioned GM transformation to a Forney-style model
is `favored' towards gauge optimization.

\begin{figure*}[t!]
\vspace{-0.05in}
 \centering
 \begin{subfigure}[b]{.245\linewidth}
    \centering
    \includegraphics[width=.99\textwidth]{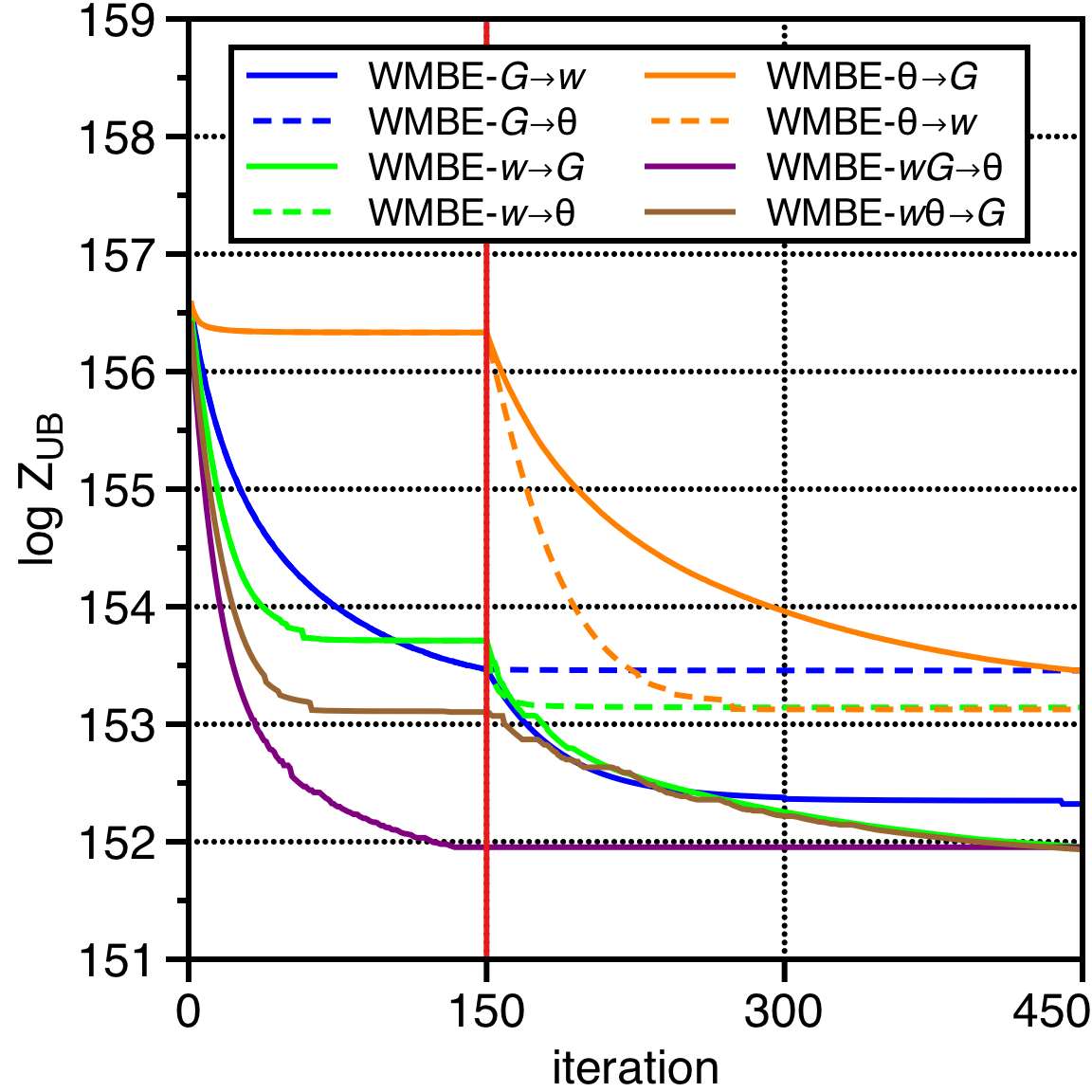}
    \caption{Ising grid GMs, $ibound=4$}
  \end{subfigure}
  \begin{subfigure}[b]{.246\linewidth}
    \centering
    \includegraphics[width=.99\textwidth]{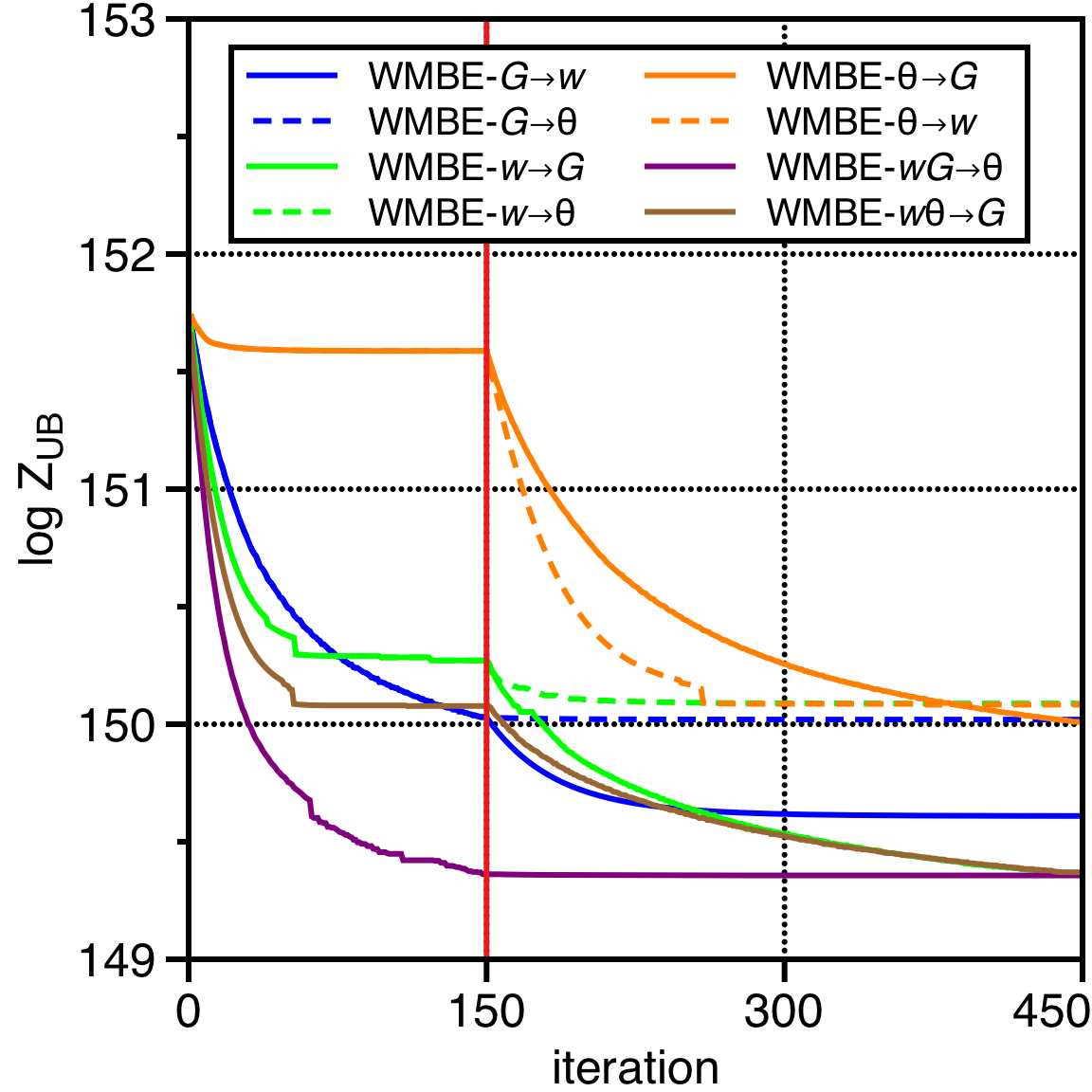}
    \caption{Ising grid GMs, $ibound=6$}
  \end{subfigure}
  \begin{subfigure}[b]{.245\linewidth}
    \centering
    \includegraphics[width=.99\textwidth]{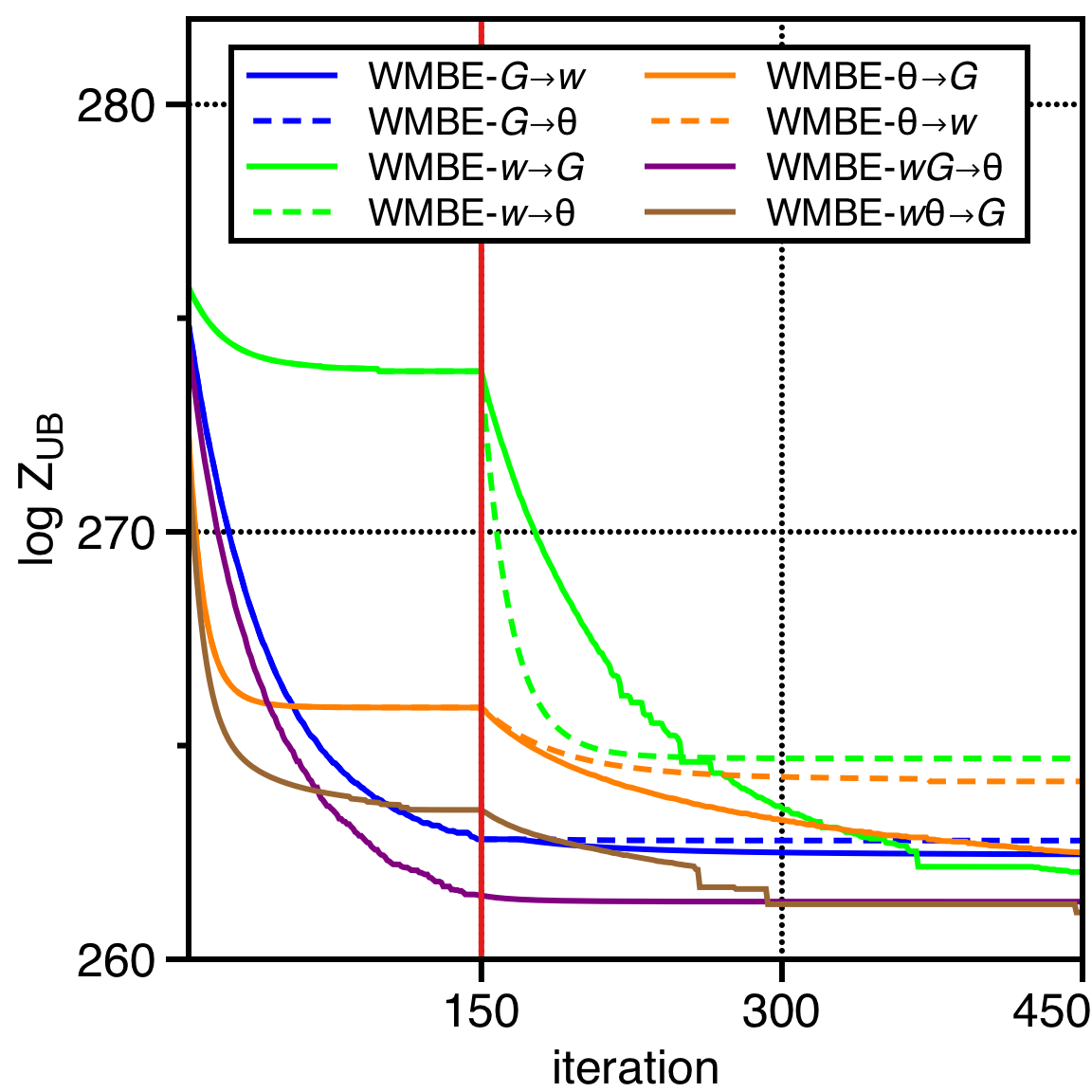}
    \caption{$3$-regular GMs, $ibound=4$}
  \end{subfigure}
  \begin{subfigure}[b]{.245\linewidth}
    \centering
    \includegraphics[width=.99\textwidth]{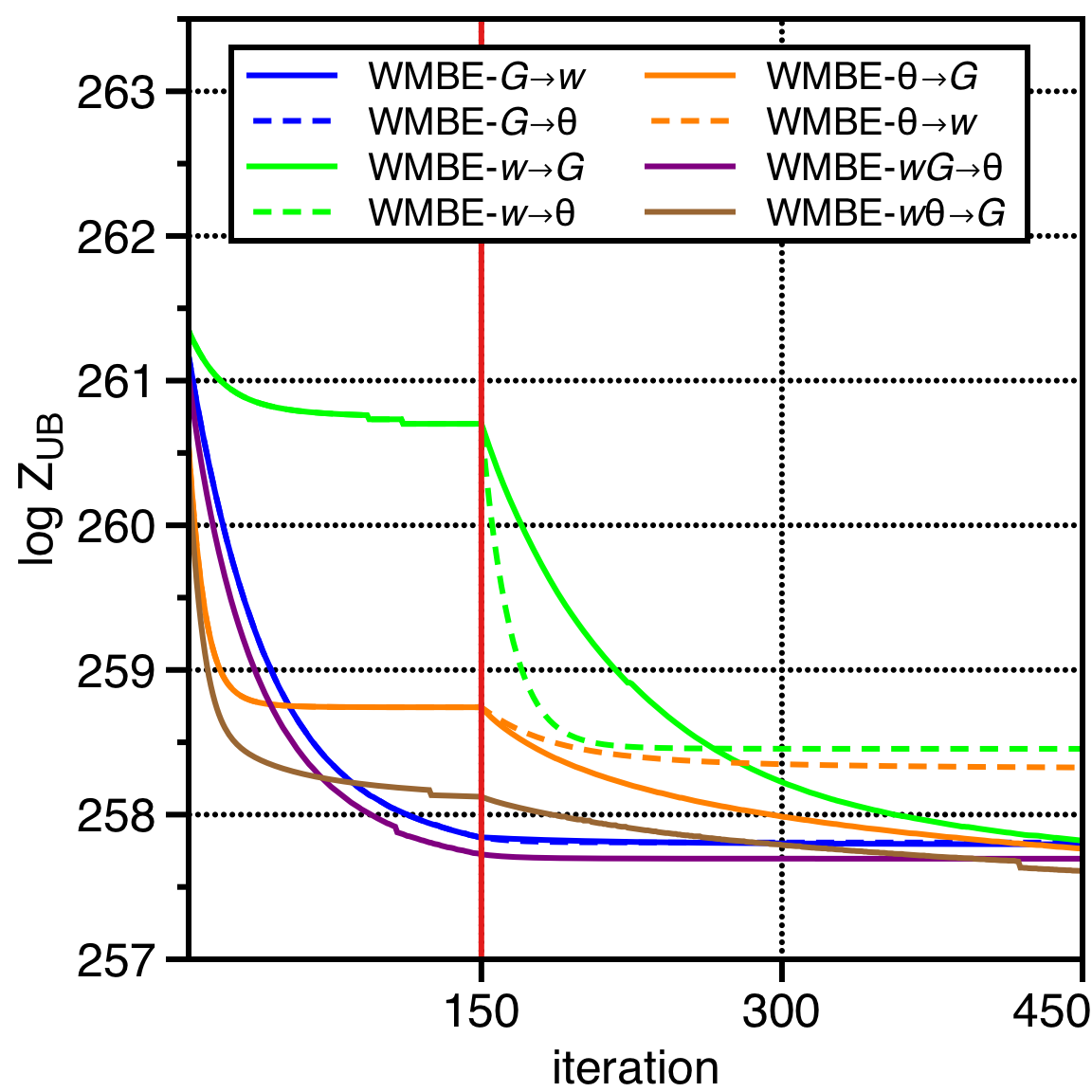}
    \caption{$3$-regular GMs, $ibound=6$}
  \end{subfigure}
  \caption{Effectiveness of optimizing various parameter choices 
  (all methods return an upper bound on $\log Z$).
  }
  \label{fig:iter}
  \vspace{-0.05in}
\end{figure*}

{\bf Further details.}
H\"older weights $w$ and reparameterizations $\theta$
were updated using projected gradients
and log-gradients respectively, as 
proposed in \cite{liu2011bounding}. 
Step sizes for gradients were chosen as $0.01, 0.1, 0.1$ for optimizing each of gauge, H\"older weights, and reparameterizations, 
respectively. 
These were chosen empirically for `easy' convergence in our experiments --
there exists room for tuning or for more sophisticated gradient descent methods such as \cite{nesterov1983method}.
TRBP was run with damping until convergence. 
For Ising grid GMs, 
we measure the log-error (with base $e$) approximating the partition function $Z$, 
i.e., $\log \left({Z_{\text{UB}}}/{Z}\right)$ where $Z_{\text{UB}}$ is 
the upper bound of a respective algorithm. 
For $3$-regular GMs, it is impossible to measure
(since $Z$ is impossible to compute),
and instead we use
the relative magnitude of bounds with respect to 
the mini-bucket upper bound $Z_{\text{MBE}}$, 
i.e., $\log \left({Z_{\text{UB}}}/{Z_{\text{MBE}}}\right)$.
Since all tested algorithms 
provide guaranteed upper bounds on $Z$, 
a lower number 
indicates better performance. 
{Further, in the UAI dataset, 
$2$ out of $17$ instances 
were omitted 
since it had factors 
with size larger than 
the algorithm's $ibound$ of our choice.}
Finally, each point in the plots represents results averaged over $10$ independent runs.

\subsection{Experimental Results}
As shown in Figures \ref{fig:performance}(a)-(b), 
TRBP and MBE perform better on the transformed Forney-style GMs than on the original 
Ising models (this may be interesting to explore in future work), but not by nearly enough to achieve the performance of the other methods. 
For fair comparison, we should examine the `TRBP' and `MBE' plots rather than the `-Ising' versions.
We observe that 
WMBE-$wG$, which enjoys the 
most  freedom in 
optimization of the WMBE bound,  
outperforms all other tested algorithms. 
In particular, 
the benefit of WMBE-$wG$ appears to increase with higher interaction strength.
Comparing optimizations of 
just one class of parameters, 
i.e., WMBE-$G$, WMBE-$w$, WMBE-$\theta$, 
we observe that 
WMBE-$G$ performs at least as well as others. 
In particular, 
optimizing gauges 
is always better than 
optimizing over the subclass of reparameterizations, i.e., 
WMBE-$G$ and WMBE-$wG$ always outperform 
WMBE-$\theta$ and WMBE-$w\theta$, respectively. 
Further, WMBE-$G$ outperforms other approaches significantly for $3$-regular GMs and UAI dataset, where 
it outperforms even 
WMBE-$w\theta$ in $3$-regular GMs 
with $ibound = 4$ and 
some instances of the 
UAI dataset.

Next, we consider experiments on
specific instances of the Ising grid GM and $3$-regular GM 
with $T=1.0$
in order to measure 
the effectiveness of 
optimizing each parameter $G$, $w$, $\theta$ separately over iterations; see Figure \ref{fig:iter}. 
Specifically, 
we first optimize a chosen parameter with respect to WMBE related bounds 
(via gradient descent methods)
for an initial $150$ iterations.
Then, we change 
the parameter to optimize further 
(e.g., $G\rightarrow \theta$) 
for another $300$ iterations 
to observe the additional benefit from optimizing the second parameter.
The running times per iteration for all parameters are comparable.
We observe that $G$ methods perform very well, which is particularly impressive since we use a small step size for gauges. 
Overall, observed performance gains may be ranked as: 
gauges $>$ weights $>$ reparameterization
for Ising grid GMs; and 
gauges $>$ reparameterization $>$ weights for 3-regular GMs.
Gauge optimization 
is critical for 
the best performance in all experiments. As expected, $wG$ yields the best results. For 3-regular GMs, gauge optimization alone is almost optimal. 

\section{Conclusion}

We developed a new gauge-variational approach to yield guaranteed bounds on the partition functions of GMs
by jointly optimizing variational parameters and gauge transformations. Our approach has better scaling characteristics then other recent state-of-the-art methods, and should be of significant practical value. 
\subsubsection*{Acknowledgements}
AW acknowledges support from the David MacKay Newton research fellowship at Darwin College, The Alan Turing Institute under EPSRC grant EP/N510129/1 \& TU/B/000074, and the Leverhulme Trust via the CFI. 
This work was partly supported by the ICT R\&D program of MSIP/IITP 
[R-20161130-004520, Research on Adaptive 
Machine Learning Technology Development 
for Intelligent Autonomous Digital Companion].
The work of MC at LANL was carried out under the auspices
of the National Nuclear Security Administration of the U.S. Department of Energy
under Contract No. DE-AC52-06NA25396.

\newpage
\bibliography{references,BP_review}

\begin{thebibliography}{10}

\bibitem{gallager1962low}
Robert Gallager.
\newblock Low-density parity-check codes.
\newblock {\em IRE Transactions on information theory}, 8(1):21--28, 1962.

\bibitem{kschischang1998iterative}
Frank~R. Kschischang and Brendan~J. Frey.
\newblock Iterative decoding of compound codes by probability propagation in
  graphical models.
\newblock {\em IEEE Journal on Selected Areas in Communications},
  16(2):219--230, 1998.

\bibitem{35Bet}
Hans~A. Bethe.
\newblock Statistical theory of superlattices.
\newblock {\em Proceedings of Royal Society of London A}, 150:552, 1935.

\bibitem{36Pei}
Rudolf~E. Peierls.
\newblock Ising's model of ferromagnetism.
\newblock {\em Proceedings of Cambridge Philosophical Society}, 32:477--481,
  1936.

\bibitem{87MPZ}
Marc M\'{e}zard, Georgio Parisi, and M.~A. Virasoro.
\newblock {\em {Spin Glass Theory and Beyond}}.
\newblock Singapore: World Scientific, 1987.

\bibitem{parisi1988statistical}
Giorgio Parisi.
\newblock Statistical field theory, 1988.

\bibitem{09MM}
Marc Mezard and Andrea Montanari.
\newblock {\em Information, Physics, and Computation}.
\newblock Oxford University Press, Inc., New York, NY, USA, 2009.

\bibitem{pearl2014probabilistic}
Judea Pearl.
\newblock {\em Probabilistic reasoning in intelligent systems: networks of
  plausible inference}.
\newblock Morgan Kaufmann, 2014.

\bibitem{jordan1998learning}
Michael~I. Jordan.
\newblock {\em Learning in graphical models}, volume~89.
\newblock Springer Science \& Business Media, 1998.

\bibitem{freeman2000learning}
William~T. Freeman, Egon~C. Pasztor, and Owen~T. Carmichael.
\newblock Learning low-level vision.
\newblock {\em International journal of computer vision}, 40(1):25--47, 2000.

\bibitem{jerrum1993polynomial}
Mark Jerrum and Alistair Sinclair.
\newblock Polynomial-time approximation algorithms for the {I}sing model.
\newblock {\em SIAM Journal on computing}, 22(5):1087--1116, 1993.

\bibitem{wainwright2005new}
Martin~J. Wainwright, Tommi~S. Jaakkola, and Alan~S. Willsky.
\newblock A new class of upper bounds on the log partition function.
\newblock {\em IEEE Transactions on Information Theory}, 51(7):2313--2335,
  2005.

\bibitem{pearl1982reverend}
Judea Pearl.
\newblock {\em Reverend Bayes on inference engines: A distributed hierarchical
  approach}.
\newblock Cognitive Systems Laboratory, School of Engineering and Applied
  Science, University of California, Los Angeles, 1982.

\bibitem{liu2010negative}
Qiang Liu and Alexander Ihler.
\newblock Negative tree reweighted belief propagation.
\newblock In {\em Proceedings of the Twenty-Sixth Conference on Uncertainty in
  Artificial Intelligence}, pages 332--339. AUAI Press, 2010.

\bibitem{ermon2012density}
Stefano Ermon, Ashish Sabharwal, Bart Selman, and Carla~P. Gomes.
\newblock Density propagation and improved bounds on the partition function.
\newblock In {\em Advances in Neural Information Processing Systems}, pages
  2762--2770, 2012.

\bibitem{liu2011bounding}
Qiang Liu and Alexander~T. Ihler.
\newblock Bounding the partition function using {H}\"older's inequality.
\newblock In {\em Proceedings of the 28th International Conference on Machine
  Learning (ICML-11)}, pages 849--856, 2011.

\bibitem{novikov2014putting}
Alexander Novikov, Anton Rodomanov, Anton Osokin, and Dmitry Vetrov.
\newblock Putting {MRF}s on a tensor train.
\newblock In {\em International Conference on Machine Learning}, pages
  811--819, 2014.

\bibitem{03WJW}
Martin~J. Wainwright, Tommy~S. Jaakkola, and Alan~S. Willsky.
\newblock Tree-based reparametrization framework for approximate estimation on
  graphs with cycles.
\newblock {\em Information Theory, IEEE Transactions on}, 49(5):1120--1146,
  2003.

\bibitem{06CCa}
Michael Chertkov and Vladimir Chernyak.
\newblock Loop calculus in statistical physics and information science.
\newblock {\em Physical Review E}, 73:065102(R), 2006.

\bibitem{06CCb}
Michael Chertkov and Vladimir Chernyak.
\newblock Loop series for discrete statistical models on graphs.
\newblock {\em Journal of Statistical Mechanics}, page P06009, 2006.

\bibitem{valiant2008holographic}
Leslie~G. Valiant.
\newblock Holographic algorithms.
\newblock {\em SIAM Journal on Computing}, 37(5):1565--1594, 2008.

\bibitem{al2011normal}
Ali Al-Bashabsheh and Yongyi Mao.
\newblock Normal factor graphs and holographic transformations.
\newblock {\em IEEE Transactions on Information Theory}, 57(2):752--763, 2011.

\bibitem{08JW}
Martin~J. Wainwright and Michael~I. Jordan.
\newblock Graphical models, exponential families, and variational inference.
\newblock {\em Foundations and Trends in Machine Learning}, 1(1):1--305, 2008.

\bibitem{forney2011partition}
G.~David Forney~Jr and Pascal~O. Vontobel.
\newblock Partition functions of normal factor graphs.
\newblock {\em arXiv preprint arXiv:1102.0316}, 2011.

\bibitem{Misha_notes}
Michael Chertkov.
\newblock Lecture notes on ``statistical inference in structured graphical
  models: Gauge transformations, belief propagation \& beyond",
  \url{https://sites.google.com/site/mchertkov/courses}, 2016.

\bibitem{ahn2017gauge}
Sung-Soo Ahn, Michael Chertkov, and Jinwoo Shin.
\newblock Gauging variational inference.
\newblock In {\em Advances in Neural Information Processing Systems 30}, pages
  2881--2890. Curran Associates, Inc., 2017.

\bibitem{wachter2006implementation}
Andreas W{\"a}chter and Lorenz~T Biegler.
\newblock On the implementation of an interior-point filter line-search
  algorithm for large-scale nonlinear programming.
\newblock {\em Mathematical programming}, 106(1):25--57, 2006.

\bibitem{dechter2003mini}
Rina Dechter and Irina Rish.
\newblock Mini-buckets: A general scheme for bounded inference.
\newblock {\em Journal of the ACM (JACM)}, 50(2):107--153, 2003.

\bibitem{forney2001codes}
G.~David Forney.
\newblock Codes on graphs: Normal realizations.
\newblock {\em IEEE Transactions on Information Theory}, 47(2):520--548, 2001.

\bibitem{levin2007tensor}
Michael Levin and Cody~P. Nave.
\newblock Tensor renormalization group approach to two-dimensional classical
  lattice models.
\newblock {\em Physical review letters}, 99(12):120601, 2007.

\bibitem{dechter1999bucket}
Rina Dechter.
\newblock Bucket elimination: A unifying framework for reasoning.
\newblock {\em Artificial Intelligence}, 113(1):41--85, 1999.

\bibitem{koller2009probabilistic}
Daphne Koller and Nir Friedman.
\newblock {\em Probabilistic graphical models: principles and techniques}.
\newblock MIT press, 2009.

\bibitem{hardy1934inequalities}
G.H. Hardy, J.E. Littlewood, and G.~P{\'o}lya.
\newblock Inequalities, 1934.

\bibitem{uai}
Vibhav Gogate.
\newblock {UAI 2014 Inference Competition}.
\newblock
  \url{http://www.hlt.utdallas.edu/~vgogate/uai14-competition/index.html},
  2014.

\bibitem{nesterov1983method}
Yurii Nesterov.
\newblock A method of solving a convex programming problem with convergence
  rate o(1/k2).
\newblock {\em Soviet Mathematics Doklady}, 27(2):372--376, 1983.

\end{thebibliography}
\bibliographystyle{unsrt}

\newpage
\appendix
\onecolumn

\begin{center}{\bf {\LARGE Supplement:}}
\end{center}

\begin{center}{\bf {\Large Gauged Mini-Bucket Elimination for Approximate Inference}}
\end{center}
\vspace{0.3in}

\section{Example of Gauge Transformations}

\begin{figure}[h!]
\centering
\begin{minipage}{1.5in}
\centering
    \includegraphics[height=1.0in]{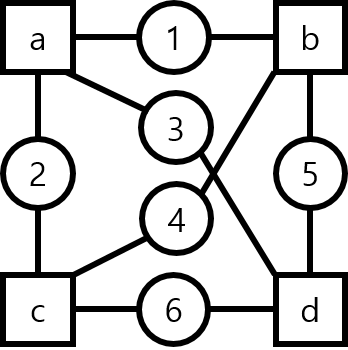}
\end{minipage}
\hspace{5pt}
\begin{minipage}{2.0in}
\hfill
{\scriptsize
\begin{align*}
    &\eff_{a}(x_{1},x_{2},x_{3}) = 
    \left[
    \left[\arraycolsep=1.0pt
    \begin{array}{cc}
         ~2432 & ~~832  \\
         ~4672 & ~~640 
    \end{array}
    \right]
    \left[\arraycolsep=1.0pt
    \begin{array}{cc}
         ~4864 & ~~384  \\
         ~5120 & ~4160 
    \end{array}
    \right]
    \right]
    \\
    &\eff_{a}(x_{1},x_{2},x_{3};\mathbf{G}_{a}) = 
    \left[
    \left[\arraycolsep=1.0pt
    \begin{array}{cc}
         ~2837 & ~1559  \\
         ~3591 & ~2077 
    \end{array}
    \right]
    \left[\arraycolsep=1.0pt
    \begin{array}{cc}
         ~3631 & ~2005  \\
         ~4261 & ~2077
    \end{array}
    \right]
    \right]
    \\
    &\eff_{b}(x_{1},x_{4},x_{5}) = 
    \left[
    \left[\arraycolsep=1.0pt
    \begin{array}{cc}
         ~1088 & ~~128  \\
         ~4928 & ~4608 
    \end{array}
    \right]
    \left[\arraycolsep=1.0pt
    \begin{array}{cc}
         ~~448 & ~1664  \\
         ~3264 & ~1344
    \end{array}
    \right]
    \right]
     \\
    &\eff_{b}(x_{1},x_{4},x_{5};\mathbf{G}_{b}) = 
    \left[
    \left[\arraycolsep=1.0pt
    \begin{array}{cc}
         ~2142 & ~1434  \\
         ~4634 & ~4558 
    \end{array}
    \right]
    \left[\arraycolsep=1.0pt
    \begin{array}{cc}
        ~~966 & ~1490 \\
        ~1490 & ~~758 
    \end{array}
    \right]
    \right]
    \\
    &\eff_{c}(x_{2},x_{4},x_{6})  = 
    \left[
    \left[\arraycolsep=1.0pt
    \begin{array}{cc}
         ~1216&~5440 \\
        ~~768&~1856 
    \end{array}
    \right]
    \left[\arraycolsep=1.0pt
    \begin{array}{cc}
         ~5568 & ~~896  \\
         ~~640 & ~~512 
    \end{array}
    \right]
    \right]
    \\
    &\eff_{c}(x_{1},x_{4},x_{5}; \mathbf{G}_{c}) = 
    \left[
    \left[\arraycolsep=1.0pt
    \begin{array}{cc}
        ~3960 &  ~8808\\
   \text{-}1608 & \text{-}2520
    \end{array}
    \right]
    \left[\arraycolsep=1.0pt
    \begin{array}{cc}
        ~\text{-}328 & \text{-}3288\\
         ~6520 &  ~8296
    \end{array}
    \right]
    \right]
    \\
    &\eff_{d}(x_{3},x_{5},x_{6}) = 
    \left[
    \left[\arraycolsep=1.0pt
    \begin{array}{cc}
         ~5632 & ~5632  \\
         ~6080 & ~6208 
    \end{array}
    \right]
    \left[\arraycolsep=1.0pt
    \begin{array}{cc}
         ~5568 & ~~896  \\
         ~~640 & ~~512 
    \end{array}
    \right]
    \right]
    \\
    &\eff_{d}(x_{3},x_{5},x_{6}; \mathbf{G}_{d}) = 
    \left[
    \left[\arraycolsep=1.0pt
    \begin{array}{cc}
       ~2408&   ~9160\\
   10760 &   ~9192
    \end{array}
    \right]
    \left[\arraycolsep=1.0pt
    \begin{array}{cc}
    14536 & \text{-}6232\\
    \text{-}7448 &  \text{-}1208
    \end{array}
    \right]
    \right]\\
&G_{1a}, G_{2a}, G_{3a}, G_{4b}, G_{5b}, G_{6c} = 
\left[\arraycolsep=1.0pt
\begin{array}{cc}
    0.75 &  0.25\\
    0.25 & 0.75
    \end{array}
\right]\\
&G_{1b}, G_{2c}, G_{3d}, G_{4c}, G_{5d}, G_{6d} = 
\left[\arraycolsep=1.0pt
\begin{array}{cc}
    1.5 &  \text{-}0.5\\
    \text{-}0.5 & 1.5
    \end{array}
\right]
\end{align*}
}
\end{minipage}
\caption{Example of gauge transformations on 
the complete graph (with respect to factors) of size $4$. 
Arrays follow row-column major indexing, e.g., 
$\eff_{a}(1,1,2) = 4864$ and $\eff_{a}(1,2,1) = 832$.}
\label{fig:my_label}
\end{figure}

\section{Proof of Theorem \ref{thm:reparamopt}} 
We prove reparameterization with respect to 
$\pmb{\theta}=\{\theta_{v\alpha}(x_{v})=0,~\forall(v,\alpha)\in E,x_{v}\}$ 
is optimal at GM with symmetric factors in the following optimization: 
\begin{align*}
    \operatornamewithlimits{\mbox{minimize}}_{\pmb{\theta}} 
    \quad
    &\wsum^{\bar{w}_{\bar{n}}}_{\bar{x}_{\bar{1}}}\cdots\wsum^{\bar{w}_{1}}_{\bar{x}_{1}}
    \prod_{\alpha \in F}\eff_{\alpha}(\bar{\mathbf{x}}_{\alpha};\pmb{\theta}_{\alpha}), \\
    \mbox{subject to}\quad
    &\prod_{\alpha \in N(v)} \exp(\theta_{v\alpha}(x_{v})) = 1
    \quad\forall~v\in X,x_{v}. 
\end{align*} 
The optimization is convex, 
and assuming $\theta_{v\beta} + \theta_{v\alpha} = 0$ from the 
constraint, 
$\partial \log Z_{\text{WMBE}}/\partial \theta_{v\alpha}=0$ 
implies optimality of the solution. 
To this end, 
the derivative 
is expressed as:
\begin{equation*}
\frac{\partial\log Z_{\text{WMBE}}}{\partial \bar{\theta}_{\alpha}(\mathbf{x}_{v\alpha})}
= \sum_{\mathbf{x}_{\alpha\backslash v}}q(\bar{\mathbf{x}}_{\alpha}) 
- \sum_{\mathbf{\bar{x}}_{\beta\backslash v}}q(\bar{\mathbf{x}}_{\beta}).
\end{equation*}
When factors are symmetric, 
it immediately follows that 
\begin{equation*}
    \sum_{\mathbf{x}_{\alpha\backslash v}}q(\bar{\mathbf{x}}_{\alpha}) = 
    \sum_{\mathbf{\bar{x}}_{\beta\backslash v}}q(\bar{\mathbf{x}}_{\beta}) 
    =0.5,
\end{equation*} 
since $q$ is expressed via 
weighted absolute sum and normalization operation of 
factors, which both 
preserve symmetry. 
Hence marginals are also symmetric, implying uniform distribution.
Hence the optimality condition is satisfied.

\end{document}